\documentclass[10pt,twocolumn,letterpaper]{article}

\usepackage{wacv}              
\usepackage[accsupp]{axessibility}  

\usepackage{graphicx}
\usepackage{amsmath}
\usepackage{amssymb}
\usepackage{booktabs}
\usepackage{amsfonts}       
\usepackage{nicefrac}       
\usepackage{microtype}      
\usepackage{xcolor}         
\usepackage{multirow}
\usepackage{ctable}
\usepackage{makecell}
\usepackage{wrapfig}
\usepackage{epsfig}
\usepackage{layouts}
\usepackage{float}
\usepackage{subcaption}

\def\mathbi#1{\textbf{\em #1}}

\usepackage[pagebackref,breaklinks,colorlinks]{hyperref}

\usepackage[capitalize]{cleveref}
\crefname{section}{Sec.}{Secs.}
\Crefname{section}{Section}{Sections}
\Crefname{table}{Table}{Tables}
\crefname{table}{Tab.}{Tabs.}

\def\wacvPaperID{1661} 
\def\confName{WACV}
\def\confYear{2024}

\begin{document}

\title{GTP-ViT: Efficient Vision Transformers via Graph-based Token Propagation}

\author{Xuwei Xu, Sen Wang, Yudong Chen\\
The University of Queensland,\\
Institution1 address\\
{\tt\small firstauthor@i1.org}
\and
Second Author\\
Institution2\\
First line of institution2 address\\
{\tt\small secondauthor@i2.org}
}

\newcommand{\inst}[1]{\textsuperscript{#1}}
\author{
Xuwei Xu\inst{1,3},\, Sen Wang\inst{1},\, Yudong Chen\inst{1,3},\, Yanping Zheng\inst{2,3},\, Zhewei Wei\inst{2},\, Jiajun Liu\inst{3,1}\\
\inst{1} The University of Queensland, Australia \\
\inst{2} Renmin University of China, China\\
\inst{3} CSIRO Data61, Australia\\
{\tt\small \{xuwei.xu, sen.wang, yudong.chen\}@uq.edu.au, \{zhengyanping, zhewei\}@ruc.edu.cn}\\
{\tt\small jiajun.liu@csiro.au} 
}

\maketitle

\begin{abstract}
Vision Transformers (ViTs) have revolutionized the field of computer vision, yet their deployments on resource-constrained devices remain challenging due to high computational demands. To expedite pre-trained ViTs, token pruning and token merging approaches have been developed, which aim at reducing the number of tokens involved in the computation. However, these methods still have some limitations, such as image information loss from pruned tokens and inefficiency in the token-matching process. In this paper, we introduce a novel \textbf{G}raph-based \textbf{T}oken \textbf{P}ropagation (\textbf{GTP}) method to resolve the challenge of balancing model efficiency and information preservation for efficient ViTs. Inspired by graph summarization algorithms, GTP meticulously propagates less significant tokens' information to spatially and semantically connected tokens that are of greater importance. Consequently, the remaining few tokens serve as a summarization of the entire token graph, allowing the method to reduce computational complexity while preserving essential information of eliminated tokens. Combined with an innovative token selection strategy, GTP can efficiently identify image tokens to be propagated. Extensive experiments have validated GTP's effectiveness, demonstrating both efficiency and performance improvements. Specifically, GTP decreases the computational complexity of both DeiT-S and DeiT-B by up to 26\% with only a minimal 0.3\% accuracy drop on ImageNet-1K \textit{without finetuning}, and remarkably surpasses the state-of-the-art token merging method on various backbones at an even faster inference speed. The source code is available at \url{https://github.com/Ackesnal/GTP-ViT}.
\noindent\begin{figure}[h]
    \setlength{\abovecaptionskip}{2pt}
    \subcaptionbox{DeiT-S as backbone \label{fig:DeiT-S-compare}}{
        \includegraphics[width=.47\columnwidth]{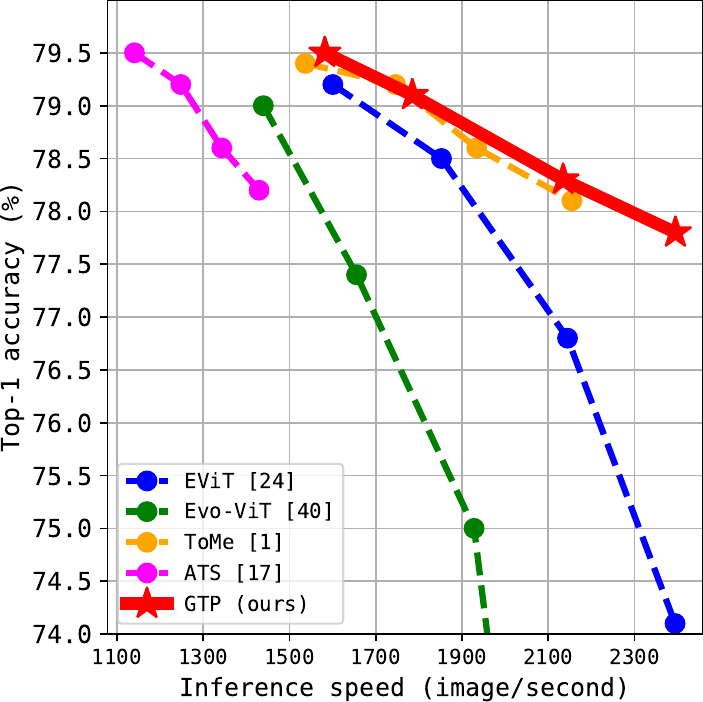}
    }
    \subcaptionbox{DeiT-B as backbone \label{fig:DeiT-B-compare}}{
        \includegraphics[width=.47\columnwidth]{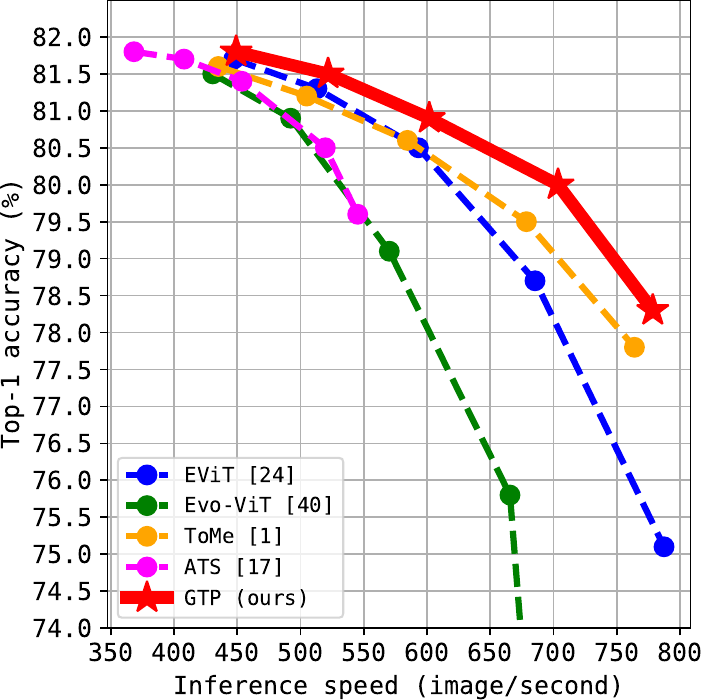}
    }
    \caption{\textbf{Comparisons among token reduction methods, taking DeiT models\cite{touvron2021training} as backbones.} Our GTP presents the best trade-off between model efficiency and performance.} 
  \label{fig:visualization tradeoff compare}
  \vspace{-1em}
\end{figure}
\end{abstract}
\begin{figure*}[t]
  \centering
  \includegraphics[width=0.9\textwidth]{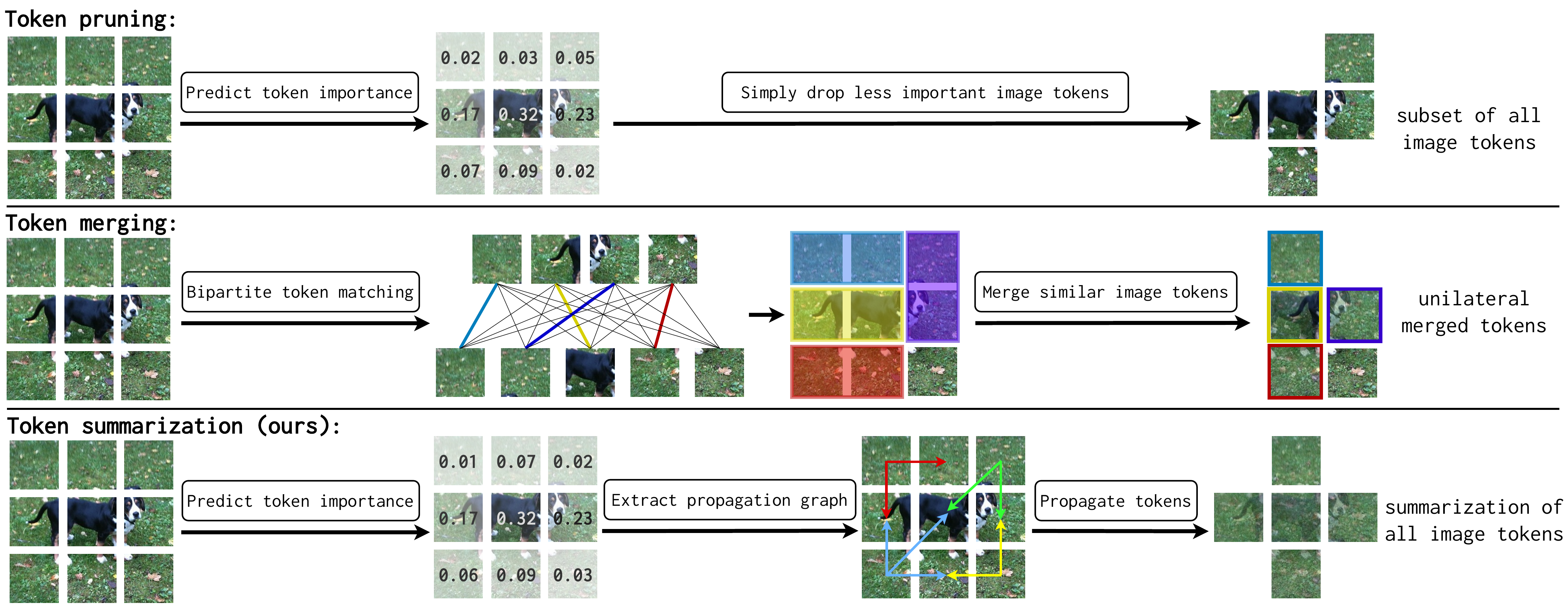}
  \vspace{-0.9em}
  \caption{\textbf{Comparisons among existing token pruning \cite{rao2021dynamicvit,liang2021evit,kong2022spvit} (top), token merging \cite{bolya2022token} (middle) and our token summarization (bottom) methods.} Both token pruning and token summarization can efficiently measure the importance of each token and determine which tokens should be discarded, providing a computational advantage over token merging. However, only token merging and token summarization successfully preserve the information of eliminated tokens.}
  \label{fig:method comparisons}
  \vspace{-0.5em}
\end{figure*}

\vspace{-0.5em}
\section{Introduction}
\vspace{-0.5em}
In recent years, Vision Transformer (ViT) has rapidly emerged as the leading backbone for various computer vision tasks, demonstrating remarkable performance in image classification \cite{dosovitskiy2020image, zhai2021scaling, ding2022davit, liu2021swin}, object detection \cite{liu2021swin, liu2022swin, chen2022vision} and segmentation \cite{chen2022vision,yang2021focal}. Despite its impressive accomplishments in the computer vision domain, the high computational cost hinders the applicability of ViT on devices with constrained computing resources. As a result, improving ViT's computational efficiency has become a growing area of interest in ViT research.

Various approaches have been explored to alleviate the heavy computational burden on ViT, such as integrating self-attention with convolution \cite{srinivas2021bottleneck, wu2021cvt, d2021convit, chen2021visformer, yoo2023enriched} and designing regional self-attention \cite{liu2021swin, chen2021crossvit, dong2021cswin}. In contrast to the methods that put forward novel efficient architectures for ViT, token pruning techniques \cite{rao2021dynamicvit,liang2021evit,marin2021token,kong2022spvit,fayyaz2022adaptive,chuanyang2022savit,xu2022evo} are proposed to expedite pre-established ViT models. In particular, token pruning methods first measure the importance of each token and then discard the insignificant ones, aiming to gradually reduce the number of tokens involved in the computation. While improving the model efficiency, pruning image tokens inevitably leads to an irreversible information loss of the removed tokens and subsequently compromises performance, especially when a large number of tokens are eliminated. Besides, token pruning methods necessitate further finetuning to prevent a significant performance drop, which also increases their computational cost. Regarding the defects of token pruning, a recent study \cite{bolya2022token} suggests token merging as a solution to preserve information and avoid finetuning. However, token merging incurs a token-matching process per layer with computational complexity proportional to the feature dimensions and the square of the number of tokens, making it less efficient than token pruning.

Limitations of the existing methods highlight a research challenge: \textbf{how to effectively balance the model efficiency and information preservation for ViT models}. Meanwhile, \textbf{how to enhance the efficiency of pre-trained ViT models while achieving minimal performance drop without finetuning} also leaves an open research question, especially in computing resource-constrained environments.

In this work, we present a novel Graph-based Token Propagation (GTP) approach to address these challenges. Motivated by graph summarization techniques \cite{lefevre2010grass, riondato2017graph, liu2018graph} that target generating condensed representations of a graph, we redefine the problem of token removal and information preservation as a token summarization task where remaining tokens encapsulate the information of the removed ones.

First, we propose an innovative and efficient token selection strategy for GTP to measure the importance of each token. The token selection strategy is based on the regeneration difficulty and broadcasting ability of each token, which can be easily drawn from the attention map that has already been calculated in the self-attention module. Consequently, our token selection strategy is more efficient than pairwise token matching in \cite{bolya2022token}, and can perform on ViTs without the [CLS] token while \cite{liang2021evit,kong2022spvit,fayyaz2022adaptive,chuanyang2022savit,xu2022evo} all depend on it. Second, inspired by the message-passing mechanism in Graph Neural Networks \cite{kipf2016semi, velivckovic2017graph, bronstein2021geometric}, GTP constructs an image token graph and distributes the information of eliminated tokens to their neighbours in the graph. As a result, GTP preserves token information through multilateral relationships among tokens while the existing token merging method \cite{bolya2022token} concentrates exclusively on one-to-one matching and merging. The kept tokens eventually make up a smaller representation of the image without abandoning much information. Figure \ref{fig:method comparisons} illustrates the comparison between our method and previous approaches. Moreover, we observe that directly discarding tokens in a pre-trained ViT tends to yield a smoother attention map after softmax activation. To resolve this issue, our approach integrates attention map sparsification as an anti-oversmoothing mechanism.

Our contributions are as follows: 1) we introduce a novel and efficient token selection strategy (Section \ref{subsubsec:token selection}); 2) we design a graph-based token propagation method to summarize the whole image, preserving information of removed tokens (Sections \ref{subsubsec:graph construction} and \ref{subsubsec:token summarization}); 3) we sparsify the attention map to enforce tokens to focus on significant information (Section \ref{subsubsec:attention sparsification}). Extensive experiments have demonstrated the effectiveness of GTP. Remarkably, taking pre-trained DeiT-B \cite{touvron2021training} as the backbone, GTP achieves 28\% real inference speed up at the cost of merely a 0.3\% accuracy drop without finetuning, and outperforms state-of-the-art token reduction methods in terms of the trade-off between performance and efficiency, as illustrated in Figure \ref{fig:visualization tradeoff compare}.

\vspace{-0.5em}
\section{Related works}
\vspace{-0.5em}
\paragraph{Efficient Vision Transformers.} Ever since the success of Vision Transformer (ViT) \cite{dosovitskiy2020image}, numerous studies have been investigating efficient ViTs. Some approaches devise fast self-attention computations that scale linearly or close to linearly with respect to either input length or feature dimensions \cite{zhang2021multi, wang2021pyramid, chu2021twins, lu2021soft, lan2023couplformer}. Besides, some combine self-attention layers with efficient convolutional layers \cite{srinivas2021bottleneck, chen2021visformer, wu2021cvt, dai2021coatnet, chen2021mobile}. Additionally, regional self-attention methods \cite{liu2021swin, dong2021cswin, chen2021regionvit, yang2021focal, chen2021dpt, liu2022swin} that calculate self-attention within a constrained area have been proposed to reduce the computation complexity of global token interactions. Distinct from these methods, our method concentrates on expediting pre-trained ViTs with plug-and-play components instead of proposing new backbone architectures.
\vspace{-1.2em}

\paragraph{Token pruning and merging.} Leveraging the inherent redundancy among image tokens, many studies have attempted to reduce the number of tokens in ViTs \cite{rao2021dynamicvit, ryoo2021tokenlearner, liang2021evit, zong2022self, fayyaz2022adaptive, xu2022evo, meng2022adavit, liu2023patchdropout}. \cite{rao2021dynamicvit} introduces a predictor module to identify tokens that can be removed without significant performance degradation. \cite{liang2021evit} removes tokens based on their attention to the [CLS] token. \cite{fayyaz2022adaptive} scores and samples important tokens to retain. However, these approaches result in information loss of the discarded tokens and necessitate finetuning from pre-trained models. Apart from the pruning-based methods, \cite{bolya2022token} proposes a token merging method that maintains the information by merging similar tokens, which does not need further finetuning. Our method builds upon these ideas by introducing a graph-based token propagation technique, addressing the limitations of existing token reduction strategies.

\begin{figure*}
    \label{fig:main architecture}
    \centering
    \includegraphics[width=0.95\textwidth]{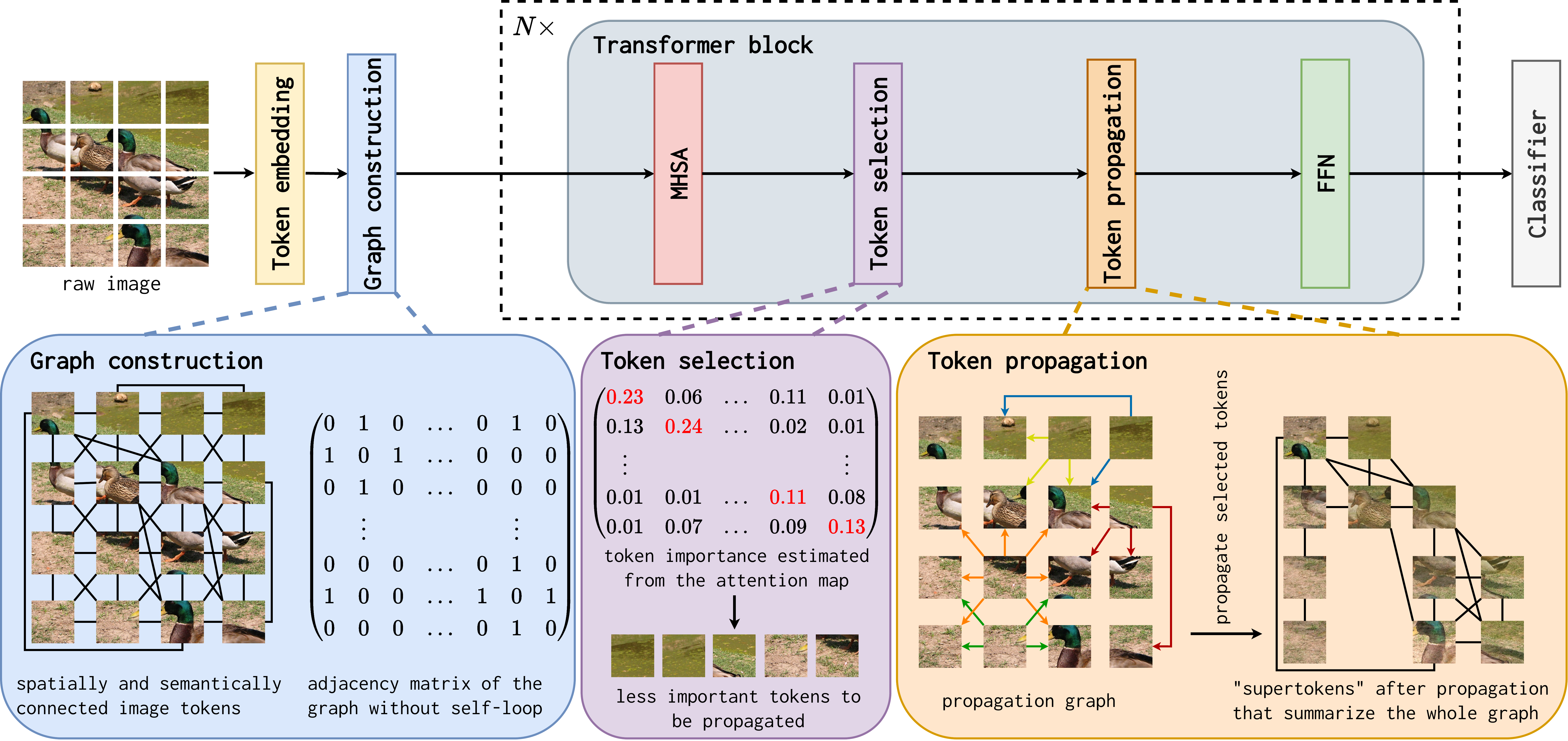}
    \vspace{-0.8em}
    \caption{\textbf{Graph-based Token Propagation (GTP) visualization.} GTP constructs a graph of image tokens after the token embedding layer \textbf{only once}. Within each transformer block, GTP utilizes the attention map computed in the MHSA layer to estimate the importance score for each image token. Next, it propagates less significant tokens to important tokens w.r.t. a subgraph that only contains edges from propagated tokens to kept tokens. As a result, the remaining tokens form a condensed graph representation of the entire image.}
\end{figure*}
\section{Methods}
\label{sec:methods}
\vspace{-0.3em}
\subsection{Preliminaries}
\label{subsec:preliminaries}
\vspace{-0.3em}
\paragraph{Vision Transformer.}
The vanilla ViT \cite{dosovitskiy2020image} divides an input image into several image patches, which are then projected into image token embeddings. We denote the embedded feature map of an image as $\mathbi{X}\in\mathbb{R}^{N \times C}$, where $N$ and $C$ are the number of tokens and the dimension of features, respectively. Each ViT block comprises a multi-head self-attention (MHSA) layer and a feed-forward network (FFN) layer.
In the MHSA layer, a layer-normalized feature map is first linearly transformed into Query ($\mathbi{Q}$), Key ($\mathbi{K}$) and Value ($\mathbi{V}$) matrices. Then, ViT calculates the similarity between each pair of tokens by the dot product between Query and Key with a softmax activation as
\begin{eqnarray}
    \abovedisplayskip=4pt
    \belowdisplayskip=4pt
    \mathbi{A} = \text{softmax}({\frac{\mathbi{Q}\mathbi{K}^\top}{\sqrt{d_\mathbi{K}}}}),
\end{eqnarray}
where $\mathbi{A}\in\mathbb{R}^{N \times N}$ is the attention map and $d_\mathbi{K}=C$ is the feature dimension of $\mathbi{K}$. Additionally, the MHSA layer calculates multiple attention maps to increase diversity. 

\vspace{-0.6em}
\paragraph{Graph Neural Network.}
Graph Neural Networks (GNNs) are typically constructed by stacking message-passing layers, during which all nodes in a graph update their representations by aggregating information from neighbours \cite{bronstein2021geometric}. This mechanism can also be regarded as each node propagating its information to the neighbouring nodes. Graph Convolutional Network (GCN) \cite{kipf2016semi} employs the convolution operation as a node aggregation method on the graph-structured data. Given a graph $\mathcal{G}=\{\mathcal{V}, \mathcal{E}\}$ that consists of a set $\mathcal{V}$ of nodes and a set $\mathcal{E}$ of edges with its adjacency matrix $\mathcal{A}\in\mathbb{R}^{|\mathcal{V}|\times|\mathcal{V}|}$, GCN updates the node feature map $\mathbi{Z}\in\mathbb{R}^{|\mathcal{V}|\times C}$ by
\begin{equation}
    \abovedisplayskip=4pt
    \belowdisplayskip=4pt
    \label{equ:gcn}
    \text{GCN}(\mathbi{Z}) = \sigma(\mathcal{D}^{-\frac{1}{2}}\mathcal{A}\mathcal{D}^{-\frac{1}{2}}\mathbi{Z}\Theta),
\end{equation}
where $\Theta$ is the projection weight, $\sigma$ represents a nonlinearity (i.e., ReLU), $\mathcal{D}^{-\frac{1}{2}}\mathcal{A}\mathcal{D}^{-\frac{1}{2}}$ is the symmetrically normalized adjacency matrix, and $\mathcal{D}$ is the degree matrix $\mathcal{D}_{i,i}=\sum_j{\mathcal{A}_{i,j}}$.

\vspace{0.3em}
\subsection{Efficient token propagation}
\label{subsec:efficient token propagation}
\vspace{-0.3em}
\subsubsection{Token selection}
\label{subsubsec:token selection}
\vspace{-0.5em}
A swift and effective token selection strategy is essential for identifying which tokens can be propagated and discarded without significantly sacrificing. In our method, we evaluate the importance of a token from two aspects.

\vspace{-1.2em}
\paragraph{Regeneration difficulty.} 
We assume that a token is less important than others if it is primarily aggregated by other tokens during the self-attention process. These less important tokens can be dropped since they are more easily to be regenerated by other tokens and their information is less significant in token summarization results. Specifically, the regeneration difficulty score $\gamma_i$ of an image token $\mathbi{x}_i$ is calculated by the negative sum of attentions from all other tokens to $\mathbi{x}_i$ as
\begin{equation}
    \abovedisplayskip=4pt
    \belowdisplayskip=4pt
    \gamma_i = \oplus\left(-\sum_{j\in\{0,\dots,N-1\}\backslash\{i\}}{\mathbi{A}_{i,j}}\right) = \oplus(\mathbi{A}_{i,i}-1),
\end{equation}
where $\oplus(\cdot)$ is a permutation-invariant aggregator to fuse the values from multiple heads. A greater $\gamma_i$ indicates a more important token $\mathbi{x}_i$. Since we only need to know the order of $\gamma$s corresponding to different image tokens rather than the values themselves, the constant term in the formula can be omitted, thereby $\gamma_i = \oplus(\mathbi{A}_{i,i})$. Consequently, the regeneration difficulty scores $\Gamma_{\mathbi{X}}$ for all the image tokens in a feature map $\mathbi{X}$ can be directly obtained from the main diagonal of its attention map $\mathbi{A}$ as $\Gamma_{\mathbi{X}}=[\gamma_0,\dots,\gamma_{N-1}]=[\oplus(\mathbi{A}_{0,0}),\dots,\oplus(\mathbi{A}_{N-1,N-1})]=diag(\oplus(\mathbi{A}))$.

\vspace{-0.8em}
\paragraph{Broadcasting ability.} 
Despite the regeneration difficulty, an image token is also indispensable if it considerably contributes to other tokens in the self-attention computation. We quantify the broadcasting ability of a token $\mathbi{x}_i$ by adding up the attention scores from this token to all other tokens and denote the score as $\psi_i$:
\begin{equation}
    \abovedisplayskip=4pt
    \belowdisplayskip=4pt
    \psi_i = \oplus\left(\sum_{j\in\{0,\dots,N-1\}\backslash\{i\}}{\mathbi{A}_{j,i}}\right).
\end{equation}
The broadcasting ability score $\psi$ reflects the significance of a token's role in broadcasting information to other tokens in ViT. Specifically, we use $\Psi_\mathbi{X}$ to denote the broadcasting abilities for all the image tokens in a feature map $\mathbi{X}$, where $\Psi_\mathbi{X}=[\psi_0,\dots,\psi_{N-1}]$.

\vspace{-0.8em}
\paragraph{Token selection.} 
Taking account of both regeneration difficulty score $\Gamma$ and broadcasting ability score $\Psi$, we keep $N-P$ tokens with the largest $\Gamma\times\Psi$ values and propagate the rest $P$ tokens. The propagated tokens are denoted by $\mathbi{X}^{p}\in\mathbb{R}^{P \times C}$ while the kept tokens are denoted by $\mathbi{X}^{k}\in\mathbb{R}^{(N-P) \times C}$, where $N$ and $P$ represent the total number of tokens and the number of propagated tokens, respectively. In addition, we exclude the [CLS] token from the token selection procedure and retain it by default. In particular, we choose $max(\cdot)$ as the $\oplus(\cdot)$ by default. 

\vspace{-1em}
\paragraph{Analysis.} 
Our token selection strategy offers three key advantages. First, in contrast to \cite{rao2021dynamicvit, kong2022spvit}, our strategy does not introduce additional parameters. Second, unlike \cite{liang2021evit}, our method can operate without the [CLS] token, which can extend our method to ViTs that do not use the [CLS] token. Moreover, our strategy is computationally efficient, as it does not necessitate the calculation of pairwise similarities among tokens, making it faster than \cite{bolya2022token} in practice. We have compared different token selection strategies in Figure \ref{fig:visualizationcompare}, including mixed strategy of both regeneration difficulty and broadcasting ability (\textit{MixedAttn}), solely regeneration difficulty (\textit{DiagAttn}), solely broadcasting ability (\textit{BroadAttn}), [CLS] token attention (\textit{CLSAttn}) \cite{liang2021evit}, cosine similarity between tokens (\textit{CosSim}) \cite{bolya2022token} and random selection (\textit{Random}).

\vspace{-1em}
\subsubsection{Sparse graph construction}
\label{subsubsec:graph construction}
\vspace{-0.4em}
GTP regards the image tokens as nodes in a graph and constructs a sparse graph based on spatial and semantic relationships between tokens. Notably, the token graph is constructed subsequent to the token embedding layer \textbf{only once} and remains static throughout the network, eliminating the need for repeated construction in each layer.

\vspace{-1em}
\paragraph{Spatial graph.} 
Since each image token corresponds to a region on the raw image, we can simply generate a spatial graph with respect to the tokens' original locations on the raw image. The adjacency matrix $\mathcal{A}^{\text{spatial}}$ of the spatial graph is defined as
\begin{equation}
    \abovedisplayskip=2pt
    \belowdisplayskip=2pt
    \mathcal{A}^{\text{spatial}}_{i,j}=\left\{\begin{array}{rl}
                                 1 &  \text{if $\mathbi{x}_i$ and $\mathbi{x}_j$ are adjacent and $i \neq j$}\\
                                 0 & \text{otherwise},
                             \end{array}
                      \right.
\end{equation}
which enables GTP to capture the spatial information of image tokens in a graph representation. The adjacency matrix $\mathcal{A}^{\text{spatial}}$ is fixed for all images.

\vspace{-1em}
\paragraph{Semantic graph.} 
While the spatial graph reflects the spatial connections among tokens, capturing their semantic connections is also essential. We leverage the cosine similarity to measure the semantic affinity between tokens $\mathbi{x}_i$ and $\mathbi{x}_j$ in the initial feature map $\mathbi{X}_0$ as
\begin{equation}
    \abovedisplayskip=2pt
    \belowdisplayskip=2pt
    \text{CosSim}(\mathbi{x}_i, \mathbi{x}_j) = \frac{\mathbi{x}_i\mathbi{x}_j^\top}{\|\mathbi{x}_i\|\cdot\|\mathbi{x}_j\|}.
\end{equation}
Then, the adjacency matrix $\mathcal{A}^{\text{semantic}}$ of the semantic graph is defined as
\begin{equation}
    \abovedisplayskip=2pt
    \belowdisplayskip=2pt
    \mathcal{A}^{\text{semantic}}_{i,j}=\left\{\begin{array}{rl}
                                 1 &  \text{if $\text{CosSim}(\mathbi{x}_i, \mathbi{x}_j)\geq T_i$ and $i \neq j$}\\
                                 0 & \text{otherwise},
                             \end{array} 
                      \right.
\end{equation}
where $T_i$ represents the $M^{\text{th}}$ ($M \ll N$) largest cosine similarity value of node $\mathbi{x}_i$ to other nodes. $T_i$ serves as a threshold to ensure that each token $\mathbi{x}_i$ has a maximum of $M$ edges. Distinct from the spatial graph, the semantic graph provides image-specific relationships for token propagation.

\vspace{-1em}
\paragraph{Mixed graph.} 
Next, we generate a mixed graph that effectively represents both the spatial and semantic relationships among tokens, by integrating the spatial graph and semantic graph. The adjacency matrix $\mathcal{A}$ of mixed graph is simply the union of $\mathcal{A}^{\text{spatial}}$ and $\mathcal{A}^{\text{semantic}}$:
\begin{equation}
    \abovedisplayskip=2pt
    \belowdisplayskip=2pt
    \mathcal{A}_{i,j}=\left\{\begin{array}{rl}
                                 1 &  \text{if $\mathcal{A}^{\text{spatial}}_{i,j}=1$ or $\mathcal{A}^{\text{semantic}}_{i,j}=1$}\\
                                 0 & \text{otherwise}.
                             \end{array}
                      \right.
\end{equation}
Note that none of the three graphs contains self-loops. In GTP, the graph structure is only used to propagate information from eliminated tokens to the remaining tokens. Therefore, a token chosen for elimination never needs to gather information from itself. Follow Equation \ref{equ:gcn}, we symmetrically normalize the graph as:
\begin{equation}
    \abovedisplayskip=2pt
    \belowdisplayskip=2pt
    \hat{\mathcal{A}}=\mathcal{D}^{-\frac{1}{2}}\mathcal{A}\mathcal{D}^{-\frac{1}{2}},
\end{equation}
where $\mathcal{D}$ is the diagonal degree matrix defined as $\mathcal{D}_{i,i}=\sum_j{\mathcal{A}_{i,j}}$. Notably, the token graph is constructed \textbf{only once} before the Transformer blocks.

\vspace{-0.8em}
\paragraph{Implementation optimizations.} 
We offer a detailed introduction to implementation optimizations for sparse graph propagation and a fast algorithm for determining the $M^{\text{th}}$ largest value in the supplementary material.
In the following experiments, we set $M=8$ unless otherwise noted. We also provide a study on the choice of $M$ in the supplementary material.
\vspace{-0.5em}

\subsubsection{Token summarization}
\vspace{-0.4em}
\label{subsubsec:token summarization}
Motivated by the message-passing mechanism in GNNs where a node distributes its information to neighbouring nodes, we put forward the token summarization process in which an image token propagates its feature to spatially and semantically connected tokens. In each layer, GTP broadcasts the propagated tokens $\mathbi{X}^p$ to the kept tokens $\mathbi{X}^k$ by 
\begin{equation}
    \abovedisplayskip=2pt
    \belowdisplayskip=2pt
    \label{eq:graph propagation}
    \mathbi{X}^s= \mathbi{X}^k + \alpha\hat{\mathcal{A}}^{p}\mathbi{X}^p,
\end{equation}
where $\alpha$ is a hyperparameter controlling the magnitude of propagated token features. The term $\hat{\mathcal{A}}^{p}\in\mathbb{R}^{(N-P)\times P}$ is extracted from the normalized adjacency matrix $\hat{\mathcal{A}}$, such that the row and column indices correspond to the kept and propagated tokens, respectively. $\mathbi{X}^s$ is the summarization of image tokens in the current layer and participates in subsequent computations. GTP implements the token propagation process immediately after the MHSA module on a layer-by-layer basis. After the token propagation procedure, we only maintain the normalized adjacency matrix $\hat{\mathcal{A}}^{s}\in\mathbb{R}^{(N-P)\times (N-P)}$ for the remaining tokens $\mathbi{X}^s$.

\vspace{-0.8em}
\subsubsection{Attention sparsification}
\label{subsubsec:attention sparsification}
\vspace{-0.4em}
\paragraph{Proportional attention.}
After reducing the number of tokens, the vanilla softmax outputs become smoother, which could negatively impact performance \cite{zhou2021deepvit, bolya2022token}. To address this issue, we introduce the \textit{proportional attention} from \cite{bolya2022token} into GTP. The proportional attention is computed as
\begin{equation}
    \abovedisplayskip=3pt
    \belowdisplayskip=3pt
    \mathbi{A} = \text{softmax}(\frac{\mathbi{Q}\mathbi{K}^\top}{\sqrt{d_\mathbi{K}}}+\log \mathbi{s}),
\end{equation}
where $\mathbi{s}\in\mathbb{R}^{N \times 1}$ represents the size of each token. Furthermore, we use $\mathbi{s}^k$ and $\mathbi{s}^p$ to denote the sizes for kept tokens and propagated tokens, respectively. The size of a kept token is dynamically updated according to the number of tokens that it has summarized:
\begin{equation}
    \abovedisplayskip=3pt
    \belowdisplayskip=1pt
    \mathbi{s}^s= \mathbi{s}^k+\alpha\hat{\mathcal{A}}^{p}\mathbi{s}^p.
\end{equation}

\vspace{-1.2em}
\paragraph{Attention map sparsification.}
In addition to proportional attention, we refine the attention map by filtering out trivial attention values. In particular, we maintain the largest $\theta N^2$ values in the attention map and assign a zero value to the rest $(1-\theta)N^2$ elements, where $N$ is the number of tokens and $\theta\in[0,1]$ represents the attention sparsity. Attention map sparsification helps to concentrate token attention on the most significant signals, thereby relieving the smoothness of the attention map and enhancing model performance.

\section{Experiments}
\label{sec:experiments}
\vspace{-0.3em}
\subsection{Implementation settings}
\label{subsec:implementation settings}
\vspace{-0.5em}
All the experiments in this section focus on the image classification task using the ImageNet-1K dataset \cite{deng2009imagenet}, which contains approximately 1.28 million training images and 50 thousand validation images. We report the top-1 accuracy on the validation set as the main performance metric. For finetuned models, we utilize the same data augmentation and training recipe as implemented in DeiT \cite{touvron2021training} and finetune for only 30 epochs. The base and minimum learning rates for finetuning are set to $10^{-5}$ and $10^{-6}$, respectively. We measure the inference speed for GTP and all other compared models on the same NVIDIA A6000 GPU with fixed batch size 128 unless noted otherwise. We ensure the PyTorch and CUDA versions are the same for all the models.
\begin{figure*}
  \centering
  \includegraphics[width=\textwidth]{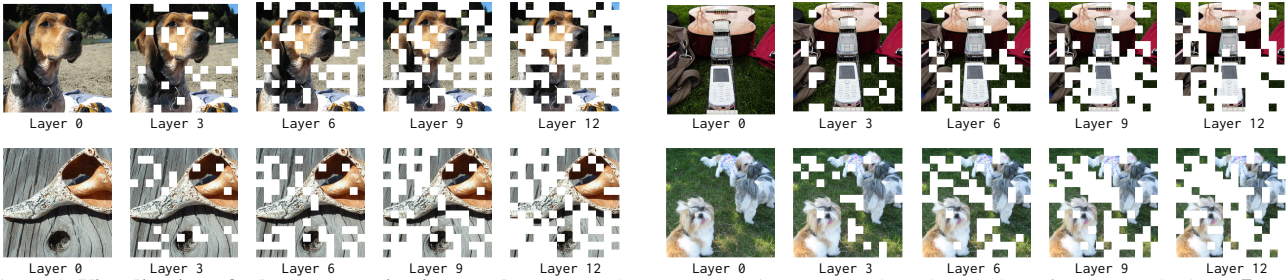}
  \vspace{-2.8em}
  \caption{\textbf{Visualization of token summarization results.} We employ GTP on DeiT-B \cite{touvron2021training} and set the number of propagated tokens $P$ to 8. Unlike existing token pruning models that focus primarily on eliminating less significant background tokens, GTP ensures the retention of certain background tokens, thereby providing a summarized representation of the original image.}
  \label{fig:visualization}
  \vspace{-0.5em}
\end{figure*}
\begin{table*}
  \scriptsize
  \setlength{\tabcolsep}{3.2pt}
  \centering
  \begin{tabular}{|l|c|c|cccccccccccccc|}
    \hline
    Backbone & \# Prop. & {\tiny $P=0$} & {\tiny $P=1$} & {\tiny $P=2$} & {\tiny $P=3$} & {\tiny $P=4$} & {\tiny $P=5$} & {\tiny $P=6$} & {\tiny $P=7$} & {\tiny $P=8$} & {\tiny $P=9$} & {\tiny $P=10$} & {\tiny $P=11$} & {\tiny $P=12$} & {\tiny $P=13$} & {\tiny $P=14$} \\
    \hline
    \multirow{3}{*}{DeiT-S \cite{touvron2021training}} & Acc. (\%) & 79.8 & 79.9 & 79.8 & 79.8 & 79.8 & 79.8 & 79.7 & 79.7 & 79.5 & 79.4 & 79.2 & 79.1 & 78.8 & 78.6 & 78.3\\
    & GMACs & 4.6 & 4.5 & 4.3 & 4.2 & 4.0 & 3.9 & 3.7 & 3.6 & 3.4 & 3.3 & 3.2 & 3.0 & 2.9 & 2.7 & 2.6\\
    & img/s & 1265.3 & 1236.7 & 1259.3 & 1325.5 & 1369.7 & 1424.2 & 1471.0 & 1511.8 & 1581.9 & 1645.0 & 1724.9 & 1784.8 & 1874.6 & 1965.5 & 2134.3\\
    \hline
    \multirow{3}{*}{DeiT-B \cite{touvron2021training}} & Acc. (\%) & 81.8 & 81.9 & 81.9 & 81.8 & 81.8 & 81.7 & 81.6 & 81.6 & 81.5 & 81.4 & 81.1 & 80.9 & 80.7 & 80.4 & 80.0\\
    & GMACs & 17.6 & 17.0 & 16.5 & 15.9 & 15.4 & 14.8 & 14.2 & 13.7 & 13.1 & 12.6 & 12.1 & 11.6 & 11.0 & 10.4 & 9.8 \\
    & img/s & 408.8 & 405.1 & 418.2 & 432.3 & 449.1 & 462.3 & 480.1 & 500.2 & 521.8 & 544.4 & 577.5 & 601.8 & 630.0 & 659.6 & 703.6 \\
    \hline
    \multirow{3}{*}{LV-ViT-S \cite{jiang2021all}} & Acc. (\%) & 83.3 & 83.0 & 82.8 & 82.6 & 82.4 & 82.4 & 82.2 & 82.1 & 81.9 & 81.7 & 81.5 & 80.8 & 80.0 & - & - \\
    & GMACs & 6.6 & 6.4 & 6.1 & 5.9 & 5.7 & 5.4 & 5.2 & 5.0 & 4.8 & 4.6 & 4.4 & 4.1 & 3.9 & - & -\\
    & img/s & 940.9 & 870.6 & 902.3 & 940.7 & 978.0 & 1038.2 & 1081.5 & 1159.7 & 1208.8 & 1258.7 & 1338.6 & 1419.2 & 1513.3 & - & -\\
    \hline
    \multirow{3}{*}{LV-ViT-M \cite{jiang2021all}} & Acc. (\%) & 84.0 & 83.8 & 83.7 & 83.6 & 83.4 & 83.3 & 83.2 & 83.0 & 82.8 & 82.5 & - & - & - & - & -\\
    & GMACs & 12.7 & 12.1 & 11.5 & 10.9 & 10.3 & 9.7 & 9.1 & 8.5 & 8.0 & 7.4 & - & - & - & - & -\\
    & img/s & 524.8 & 496.9 & 521.3 & 549.9 & 582.5 & 624.3 & 667.4 & 715.7 & 770.6 & 833.6 & - & - & - & - & -\\
    \hline
  \end{tabular}
  \vspace{-1.2em}
  \caption{\small\textbf{GTP main results on ImageNet-1K \textit{without finetuning.}} In this table, we report the best top-1 accuracy for various numbers of propagated tokens $P$ among different hyperparameter settings. $P=0$ represents the full-size backbone model. Note that LV-ViT-S and LV-ViT-M \cite{jiang2021all} can reduce at most 12 and 9 tokens per layer, respectively.}
  \label{tab:main results}
  \vspace{-0.6em}
\end{table*}

\vspace{-0.3em}
\subsection{Main result}
\vspace{-0.5em}
\label{subsec:main results}
We first apply our GTP on pre-trained DeiT-S \cite{touvron2021training}, DeiT-B \cite{touvron2021training}, LV-ViT-S \cite{jiang2021all} and LV-ViT-B \cite{jiang2021all} without additional finetuning and present the performance for various numbers of propagated tokens in Table \ref{tab:main results}. These four models are popular ViT backbones for token reduction methods. Table \ref{tab:main results} demonstrates the capability of GTP to expedite ViT without necessitating finetuning. In particular, when propagating 8 tokens per layer (i.e., $P=8$), GTP achieves 25\% real throughput speed-up (1581.3 image/s vs 1268.3 image/s) and 26\% fewer computational complexity (3.4 GMACs vs 4.6 GMACs) with an insignificant accuracy decrease of 0.3\% (79.5\% vs 79.8\%) compared to full-size DeiT-S model. Even for the more complex model, DeiT-B, GTP still accomplishes a 26\% reduction in computational complexity (13.1GMACs vs 17.6GMACs) and approximately 28\% improvement in inference speed (521.8 image/s vs 408.8 image/s) with a mere 0.3\% drop in accuracy (81.5\% vs 81.8\%). We also visualize some token summarization examples in Figure \ref{fig:visualization}.

\vspace{-0.3em}
\subsection{Comparisons with state-of-the-art methods}
\vspace{-0.5em}
\label{subsec:main comparisons}
In Tables \ref{tab:comparisons deit-s} and \ref{tab:comparisons deit-b}, we present comparisons of GTP against token pruning and token merging methods, including DynamicViT \cite{rao2021dynamicvit}, EViT \cite{liang2021evit}, ATS \cite{fayyaz2022adaptive}, Evo-ViT \cite{xu2022evo}, Tri-Level \cite{kong2022peeling} and ToMe \cite{bolya2022token}. We present the top-1 accuracy, computational complexity (measured in GMACs), and inference speed (measured in images per second) for comparison. We compare these benchmarks since they have released their official source codes so that we can reproduce the results for these models both with and without finetuning for various computational complexities. More implementation details are provided in the table captions.
\begin{table*}
  \centering
  \scriptsize
  \setlength{\tabcolsep}{1.9pt}
  \begin{tabular}{|l|l|ccc|ccc|ccc|ccc|}
    \hline
    \multirow{3}{*}{Method} & \multirow{3}{*}{\#Param} & \multicolumn{3}{c|}{Approx. 3.5 GMACs} & \multicolumn{3}{c|}{Approx. 3.0 GMACs} & \multicolumn{3}{c|}{Approx. 2.6 GMACs} & \multicolumn{3}{c|}{Approx. 2.3 GMACs} \\
    \cline{3-14}
    & & \makecell{w/ F  \\ Acc (\%)} & \makecell{w/o F\\ Acc (\%)} & \makecell{Speed\\(img/s)} & \makecell{w/ F \\ Acc (\%)} & \makecell{w/o F \\ Acc (\%)} & \makecell{Speed\\(img/s)} & \makecell{w/ F \\ Acc (\%)} & \makecell{w/o F\\ Acc (\%)} & \makecell{Speed\\(img/s)} & \makecell{w/ F \\ Acc (\%)} & \makecell{w/o F \\ Acc (\%)} & \makecell{Speed\\(img/s)}\\
    \hline
    DyViT \cite{rao2021dynamicvit} & 22.8M & 79.6 & 74.0 & 1478.4 ($\times 1.30$) & 79.3 & 67.4 & 1700.1 ($\times 1.36$) & 78.5 & 58.3 & 1980.9 ($\times 1.47$) & 77.5 & 51.7 & 2217.7 ($\times 1.48$)\\
    EViT \cite{liang2021evit} & \textbf{22.1M} & \textbf{79.8} & 79.2 & \textbf{1600.5 ($\mathbf{\times 1.40}$)} & 79.5 & 78.5 & \textbf{1852.3} ($\mathbf{\times 1.48}$) & 78.9 & 76.8 & \textbf{2144.5} ($\mathbf{\times 1.60}$) & 78.5 & 74.1 & \textbf{2393.8} ($\mathbf{\times 1.67}$)\\
    Evo-ViT \cite{xu2022evo} & \textbf{22.1M} & 78.4 & 79.0 & 1439.3 ($\times 1.26$) & 78.2 & 77.4 & 1655.4 ($\times 1.33$) & 78.0 & 75.0 & 1927.9 ($\times 1.44$) & 77.7 & 72.1 & 2016.7 ($\times 1.41$)\\
    Tri-Level \cite{kong2022peeling} & \textbf{22.1M} & 79.5 & 67.6 & 1345.9 ($\times 1.08$) & 79.1 & 67.6 & 1551.2 ($\times 1.24$) & 78.8 & 67.6 & 1793.8 ($\times 1.34$) & 78.1 & 67.6 & 2013.2 ($\times 1.41$)\\
    ToMe \cite{bolya2022token} & \textbf{22.1M} & 79.7 & 79.4 & 1536.7 ($\times 1.35$) & 79.4 & \textbf{79.2} & 1746.0 ($\times 1.40$) & 78.9 & \textbf{78.6} & 1934.0 ($\times 1.44$) & 78.4 & 78.1 & 2154.7 ($\times 1.51$)\\
    ATS \cite{fayyaz2022adaptive} & \textbf{22.1M} & 79.7 & \textbf{79.5} & 1140.6 ($\times 1.00$)& \textbf{79.7} & \textbf{79.2} & 1248.1 ($\times 1.00$)& 79.0 & \textbf{78.6} & 1343.0 ($\times 1.00$) & \textbf{78.6} & \textbf{78.2} & 1429.4 ($\times 1.00$)\\
    GTP (ours) & \textbf{22.1M} & 79.7 & \textbf{79.5} & 1581.9 ($\times 1.39$) & 79.5 & 79.1 & 1784.8 ($\times 1.43$) & \textbf{79.1} & 78.3 & 2134.3 ($\times 1.59$) & 
    \textbf{78.6} & 77.8 & 2304.7 ($\times 1.61$)\\
    \hline
  \end{tabular}
  \vspace{-1.2em}
  \caption{\small\textbf{Comparisons with state-of-the-art methods, taking DeiT-S \cite{touvron2021training} as the backbone.} "w/ F" and "w/o F" represent the performance with and without 30-epoch finetuning, respectively. We categorize the performance in terms of computational complexity. For example, \textit{Approx. 3.5GMACs} stands for the computational complexity at about 3.5 GMACs, which is equivalent to the keep ratio at 0.8 for DynamicViT \cite{rao2021dynamicvit}, EViT \cite{liang2021evit} and Tri-Level \cite{kong2022peeling}, selection ratio at 0.7 for Evo-ViT \cite{xu2022evo}, ATS block from layers 7 to 11 for ATS \cite{fayyaz2022adaptive} and the number of reduced tokens at 8 per layer for ToMe \cite{bolya2022token} and our GTP. More details are provided in the supplementary material. We leverage the slowest inference speed in each category as the baseline (i.e. $\times 1.00$) for speed comparisons. To ensure fairness, we reproduce the finetuned results for these models using their officially released codes. Figure \ref{fig:DeiT-S-compare} shows visualized comparisons. Bold font means better.}
  \label{tab:comparisons deit-s}
  \vspace{-0.3em}
\end{table*}
\begin{table*}
  \centering
  \scriptsize
  \setlength{\tabcolsep}{3pt}
  \begin{tabular}{|l|c|cc|cc|cc|cc|cc|}
    \hline
    \multirow{2}{*}{Method} & \multirow{2}{*}{\#Param} & \multicolumn{2}{c|}{Approx. 15.3 GMACs} & \multicolumn{2}{c|}{Approx. 13.1 GMACs} & \multicolumn{2}{c|}{Approx. 11.6 GMACs} & \multicolumn{2}{c|}{Approx. 9.8 GMACs} & \multicolumn{2}{c|}{Approx. 8.8 GMACs}\\
    \cline{3-12}
     & & Acc (\%) &  Speed (img/s) & Acc (\%) & Speed (img/s) & Acc (\%) & Speed (img/s) & Acc (\%)  & Speed (img/s) & Acc (\%) & Speed (img/s) \\
    \hline
    DyViT \cite{rao2021dynamicvit} & 89.5M & 79.9 & 420.0 ($\times 1.14$) & 77.7 & 489.0 ($\times 1.20$) & 75.5 & 579.7 ($\times 1.28$) & 69.5 & 676.2 ($\times 1.30$) & 50.2 & 784.2 ($\times 1.44$)\\
    EViT \cite{liang2021evit} & \textbf{86.6M} & 81.7 & 447.2 ($\times 1.21$) & 81.3 & 513.0 ($\times 1.26$) & 80.5 & 593.2 ($\times 1.31$) & 78.7 & 685.4 ($\times 1.32$) & 75.1 & \textbf{787.4 ($\mathbf{\times 1.44}$)}\\
    Evo-ViT \cite{xu2022evo} & \textbf{86.6M} & 81.5 & 430.6 ($\times 1.17$) & 80.9 & 492.2 ($\times 1.21$) & 79.1 & 570.2 ($\times 1.26$) & 75.8 & 665.6 ($\times 1.28$) & 60.6 & 736.6 ($\times 1.35$)\\
    Tri-Level \cite{kong2022peeling} & \textbf{86.6M} & 64.6 & 398.9 ($\times 1.08$) & 64.6 & 466.7 ($\times 1.14$) & 64.6 & 539.7 ($\times 1.19$) & 64.6 & 622.9 ($\times 1.20$) & 64.6 & 707.7 ($\times 1.30$)\\
    ToMe \cite{bolya2022token} & \textbf{86.6M} & 81.6 & 435.1 ($\times 1.18$) & 81.2 & 504.9 ($\times 1.24$) & 80.6 & 584.5 ($\times 1.29$) & 79.5 & 678.6 ($\times 1.31$) & 77.8 & 764.0 ($\times 1.40$)\\
    ATS \cite{fayyaz2022adaptive} & \textbf{86.6M} & \textbf{81.8} & 368.3 ($\times 1.00$) & \textbf{81.7} & 407.9 ($\times 1.00$) & \textbf{81.4} & 453.7 ($\times 1.00$) & \textbf{80.5} & 519.6 ($\times 1.00$) & \textbf{79.6} & 545.2 ($\times 1.00$)\\
    GTP (ours) & \textbf{86.6M} & \textbf{81.8} & \textbf{449.1 ($\mathbf{\times 1.22}$)} &  81.5 & \textbf{521.8 ($\mathbf{\times 1.28}$)} & 80.9 & \textbf{601.8 ($\mathbf{\times 1.33}$)} & 80.0 & \textbf{703.6 ($\mathbf{\times 1.35}$)} & 78.3 & 778.5 ($\times 1.43$)\\
    \hline
  \end{tabular}
  \vspace{-1.2em}
  \caption{\small\textbf{Comparisons with state-of-the-art methods \textit{without finetuning}, taking DeiT-B \cite{touvron2021training} as the backbone.} Due to the computing resource limitation, we only present the finetune-free results for DeiT-B. All the formats presented in this table align with the format in Table~\ref{tab:comparisons deit-s}. Figure \ref{fig:DeiT-B-compare} shows visualized comparisons. Bold font means better.}
  \label{tab:comparisons deit-b}
  \vspace{-0.8em}
\end{table*}
\noindent\begin{figure}
    \setlength{\abovecaptionskip}{3pt}
    \subcaptionbox{DeiT-S \label{fig:DeiT-S}}{
        \vspace{-0.25em}
        \includegraphics[width=.46\columnwidth]{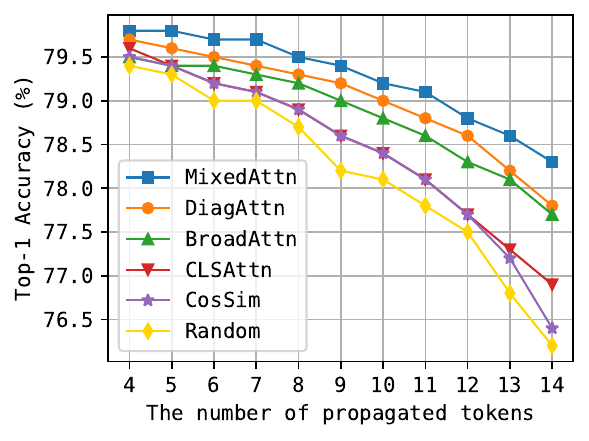}
    }
    \hspace{0.02\columnwidth}
    \subcaptionbox{DeiT-B \label{fig:DeiT-B}}{
        \vspace{-0.25em}
        \includegraphics[width=.46\columnwidth]{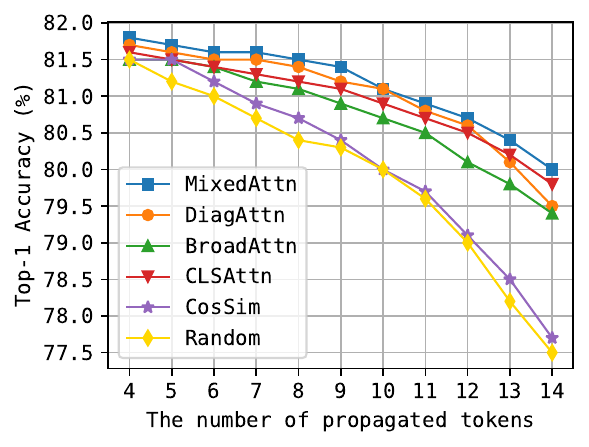}
    }
   \vspace{-0.4em}
   \caption{\textbf{Comparisons of different token selection strategies.} We apply different token selection strategies with GTP and report the top-1 accuracy for various numbers of propagated tokens ($P$).} 
   \label{fig:visualizationcompare}
   \vspace{-0.7em}
\end{figure}

Table \ref{tab:comparisons deit-s} displays both finetuned and finetune-free results on DeiT-S. At a similar inference speed, GTP can match the performance of token pruning methods with finetuning and outperform them when a larger number of tokens are eliminated. For instance, at the same computational complexity of 2.6GMACs, GTP exceeds EViT by 0.2\% top-1 accuracy (79.1\% vs 78.9\%), which reflects GTP's ability to preserve information. It is worth noting that when a substantial number of tokens are dropped, token pruning methods would suffer a dramatic accuracy drop without finetuning. Table \ref{tab:comparisons deit-b} presents the finetune-free results on DeiT-B, where GTP surpasses all the compared models at a similar inference speed. The naive elimination of tokens leads to a considerable performance decline in token pruning methods. For instance, when reducing the computational complexity of DeiT-B to 8.8GMACs, EViT can only obtain 75.1\% top-1 accuracy, which is worse than its performance on DeiT-S (76.8\%) with merely 2.6GMACs complexity. On the contrary, GTP reaches 78.3\% accuracy, surpassing EViT by a significant 3.2\% top-1 accuracy. The only exception, ATS, maintains the performance at the cost of extremely slower inference speed. This indicates that token pruning methods struggle to preserve the information of pruned tokens while remaining efficient and are ineffective without finetuning. In contrast, GTP achieves the best trade-off between model performance and efficiency without finetuning, which is also shown in Figure \ref{fig:visualization tradeoff compare}.

\subsection{Ablation studies}
\vspace{-0.5em}
\label{subsec:ablation studies}
\subsubsection{Token selection strategy}
\vspace{-0.7em}
\label{subsubsec:ablation on token selection}
In Figure \ref{fig:visualizationcompare}, we compare different token selection strategies, including mixed strategy of both regeneration difficulty and broadcasting ability (\textit{MixedAttn}), solely regeneration difficulty (\textit{DiagAttn}), solely broadcasting ability (\textit{BroadAttn}), [CLS] token attention (\textit{CLSAttn}) \cite{liang2021evit}, cosine similarity between tokens (\textit{CosSim}) \cite{bolya2022token} and random selection (\textit{Random}). Figure \ref{fig:visualizationcompare} demonstrates that our token selection strategy consistently outperforms the other methods w.r.t. different numbers of eliminated tokens. Besides, simply adopting the regeneration difficulty score (i.e., \textit{DiagAttn}) can achieve close or even higher performance than that of the traditional [CLS] attention. Considering that many new ViT architectures do not contain the [CLS] token, our method illustrates a potential solution for token pruning methods on them.

\vspace{-0.7em}
\subsubsection{Graph type}
\vspace{-0.7em}
\label{subsubsec:ablation on graph type}
We study the token graph types in the token summarization process and report their best top-1 accuracy in Table \ref{tab:graph type}. Graph type \textit{None} represents for only selecting and removing tokens without propagation, which is analogous to layer-by-layer token pruning. For semantic graphs, we set the number of semantically connected tokens ($M$) to 8 by default, thereby creating a graph of equivalent size to the spatial graph.

Primarily, we note that relying solely on a semantic graph can yield substantial performance as using the mixed graph in most scenarios. It indicates that the semantic relationship among tokens is more important than the spatial relationship.
\begin{table}[H]
    \centering
    \scriptsize
    \setlength{\tabcolsep}{4pt}
    \begin{tabular}{|l|l|cccc|}
        \hline
        \multirow{2}{*}{Backbone} & \multirow{2}{*}{Graph type} & \multicolumn{4}{c|}{Acc (\%)} \\
        \cline{3-6}
        & & {\tiny $P=4$} & {\tiny $P=8$} & {\tiny $P=11$} & {\tiny $P=14$} \\
        \hline
        \multirow{4}{*}{DeiT-S \cite{touvron2021training}}
        & Mixed & \textbf{79.8} & \textbf{79.5} & \textbf{79.1} & \textbf{78.3} \\ 
        & Spatial & \textbf{79.8} & \textbf{79.5} & 78.9 & 78.1 \\
        & Semantic & \textbf{79.8} & 79.4 & 79.0 & 78.2 \\
        & None & 79.6 & 79.2 & 78.7 & 77.6\\
        \hline
        \multirow{4}{*}{DeiT-B \cite{touvron2021training}}
        & Mixed & 81.7 & \textbf{81.5} & \textbf{80.9} & \textbf{80.0} \\
        & Spatial & \textbf{81.8} & 81.4 & \textbf{80.9} & 79.8 \\
        & Semantic & \textbf{81.8} & \textbf{81.5} & \textbf{80.9} & \textbf{80.0} \\
        & None & 81.7 & 81.4 & 80.7 & 79.6 \\
        \hline
    \end{tabular}
    \vspace{-1.2em}
    \caption{\textbf{Ablation study on graph types.}}
    \label{tab:graph type}
\end{table}
\vspace{-1.5em}
\begin{table}[H]
    \centering
    \scriptsize
    \setlength{\tabcolsep}{4pt}
    \begin{tabular}{|l|c|c|cccc|}
        \hline
        \multirow{2}{*}{Backbone} & \multirow{2}{*}{Sparsity} & Prop. & \multicolumn{4}{c|}{Acc (\%)} \\
        \cline{4-7}
        & & attn. & {\tiny $P=4$} & {\tiny $P=8$} & {\tiny $P=11$} & {\tiny $P=14$} \\
        \hline
        \multirow{7}{*}{DeiT-S \cite{touvron2021training}}
        & 1.0 & n/a & 79.7 & 79.2 & 78.7 & 77.6\\ 
        \cline{2-7}
        & 0.9 & \multirow{5}{*}{$\surd$}  & 79.7 & 79.3 & 79.0 & 78.2\\
        & 0.8 & & 79.7 & 79.4 & 79.0 & 78.1\\
        & 0.7 & & \textbf{79.8} & 79.4 & \textbf{79.1} & 78.2\\
        & 0.6 & & \textbf{79.8} & \textbf{79.5} & 79.0 & \textbf{78.3}\\
        & 0.5 & & \textbf{79.8} & 79.4 & 79.0 & 78.0\\
        \cline{2-7}
        & best & $\times$ & 79.7 & 79.3 & 78.8 & 77.8\\
        \hline
        \multirow{7}{*}{DeiT-B \cite{touvron2021training}}
        & 1.0 & n/a & 81.7 & 81.4 & 80.7 & 79.8 \\ 
        \cline{2-7}
        & 0.9 & \multirow{5}{*}{$\surd$}  & 81.7 & 81.4 & \textbf{80.9} & \textbf{80.0}\\
        & 0.8 & & 81.7 & \textbf{81.5} & \textbf{80.9} & 79.9 \\
        & 0.7 & & 81.7 & \textbf{81.5} & \textbf{80.9} & 79.9\\
        & 0.6 & & \textbf{81.8} & 81.4 & \textbf{80.9} & 79.9\\
        & 0.5 & & \textbf{81.8} & \textbf{81.5} & \textbf{80.9} & \textbf{80.0} \\
        \cline{2-7}
        & best & $\times$ & \textbf{81.8} & \textbf{81.5} & \textbf{80.9} & 79.9 \\
        \hline
    \end{tabular}
    \vspace{-1em}
    \caption{\textbf{Ablation studies on attention sparsity.} Attention sparsity at 1.0 represents the result without the attention sparsification. \textit{Prop. attn.} stands for using proportional attention. \textit{Best} stands for the best attention sparsity for each number of propagated tokens $P$.}
    \label{tab:attention sparsity}
\end{table}
\vspace{-0.3em}
\begin{table*}
  \scriptsize
  \setlength{\tabcolsep}{2.8pt}
  \centering
  \begin{tabular}{|l|cc|cc|cc|cc|cc|cc|}
    \hline
    \multirow{2}{*}{Method} & \multicolumn{2}{c|}{$P=0$} & \multicolumn{2}{c|}{$P=10$} & \multicolumn{2}{c|}{$P=20$} & \multicolumn{2}{c|}{$P=30$} & \multicolumn{2}{c|}{$P=40$} & \multicolumn{2}{c|}{$P=50$} \\
    \cline{2-13}
    & Acc (\%) & Speed (img/s) & Acc (\%) & Speed (img/s) & Acc (\%) & Speed (img/s) & Acc (\%) & Speed (img/s) & Acc (\%) & Speed (img/s) & Acc (\%) & Speed (img/s)\\
    \hline
    ToMe \cite{bolya2022token} & \multirow{2}{*}{85.8} & \multirow{2}{*}{48.3} & 85.8 & 53.5 & 85.7 & 59.7 & 85.5 & 67.2 & 85.3 & 75.9 & 84.9 & 86.8\\
    GTP (ours) & & & \textbf{85.9} & \textbf{57.6 (+7.6\%)} & \textbf{85.8} & \textbf{63.7 (+6.7\%)} & \textbf{85.7} & \textbf{74.6  (+11.0\%)} & \textbf{85.5} & \textbf{83.2 (+9.6\%)} & \textbf{85.3} & \textbf{95.5 (+10.0\%)}\\
    \hline
  \end{tabular}
  \vspace{-1.2em}
  \caption{\small\textbf{GTP performance on ViT with more tokens.} We validate GTP's effectiveness and efficiency on ViT-B-Patch8 \cite{dosovitskiy2020image}, which has 785 tokens. Due to our GPU memory constraints, the inference speeds are tested with batch size 32.}
  \label{tab:more token results}
  \vspace{-0.2em}
\end{table*}
\begin{table*}
  \scriptsize
  \vspace{-0.8em}
  \setlength{\tabcolsep}{3.2pt}
  \centering
  \begin{tabular}{|c|c|c|cccccccccccccc|}
    \hline
    Backbone & \# Prop. & {\tiny $P=0$} & {\tiny $P=1$} & {\tiny $P=2$} & {\tiny $P=3$} & {\tiny $P=4$} & {\tiny $P=5$} & {\tiny $P=6$} & {\tiny $P=7$} & {\tiny $P=8$} & {\tiny $P=9$} & {\tiny $P=10$} & {\tiny $P=11$} & {\tiny $P=12$} & {\tiny $P=13$} & {\tiny $P=14$} \\
    \hline
    \multirow{3}{*}{ViT-Medium-GAP \cite{dosovitskiy2020image}} & Acc. (\%) & 84.0 & 84.1 & 84.0 & 83.8 & 83.8 & 83.7 & 83.7 & 83.7 & 83.6 & 83.5 & 83.4 & 83.3 & 83.1 & 82.7 & 82.3\\
    & GMACs & 10.6 & 10.4 & 10.1 & 9.9 & 9.6 & 9.3 & 9.0 & 8.8 & 8.5 & 8.3 & 8.0 & 7.8 & 7.5 & 7.2 & 7.0\\
    & img/s & 535.0 & 523.7 & 536.8 & 552.7 & 566.7 & 581.5 & 598.0 & 615.3 & 635.0 & 656.5 & 670.8 & 748.7 & 778.9 & 805.6 & 833.6\\
    \hline
  \end{tabular}
  \vspace{-1.2em}
  \caption{\small\textbf{GTP results on ViT model without the [CLS] token \textit{without finetuning}.}}
  \label{tab:noclstoken}
  \vspace{-1em}
\end{table*}
\noindent Secondly, we observe that the effectiveness of graph propagation signifies when the number of eliminated tokens $P$ increases. For instance, on DeiT-B, graph propagation can only enhance the top-1 accuracy by 0.1\% when $P=4$ or $P=8$, but this accuracy difference escalates to 0.4\% when $P=14$. Meanwhile, graph propagation benefits small models more than large models. For example, graph propagation can increase the top-1 accuracy by 0.7\% on DeiT-S when $P=14$, in contrast to only 0.4\% for DeiT-B, which is a significant improvement in the field of expediting ViTs. We think this is because larger ViT models are more robust when confronted with image information loss.

\vspace{-1.1em}
\subsubsection{Attention sparsification}
\vspace{-0.7em} 
We conduct an ablation of the proportional attention utilized in the GTP method, as detailed in Table \ref{tab:attention sparsity}. For DeiT-S, we observe that proportional attention consistently enhances performance. For example, it increases the best top-1 accuracy of DeiT-S by 0.5\% when $P=14$. However, it becomes less effective on DeiT-B with only trivial improvements. We then explore the attention sparsity presenting the top-1 accuracy for different attention sparsities. Similar to the results of proportional attention, we find that attention sparsification is less impactful for larger models. For example, when removing 14 tokens per layer, a proper attention sparsity can increase the top-1 accuracy for DeiT-S by 0.7\%, much higher than the accuracy increase on DeiT-B, which is only 0.2\%. We think this is also because larger models are more robust than their smaller counterparts and already concentrate on the most significant parts of an image. In other words, the attention map of DeiT-B is already fairly sparse.

\vspace{-0.3em}
\subsection{Anti-oversmoothing}
\label{subsec:anti-oversmoothing}
\begin{figure}
    \setlength{\abovecaptionskip}{3pt}
    \subcaptionbox{Approx. 2.6 GMACs}{
        \includegraphics[width=.46\columnwidth]{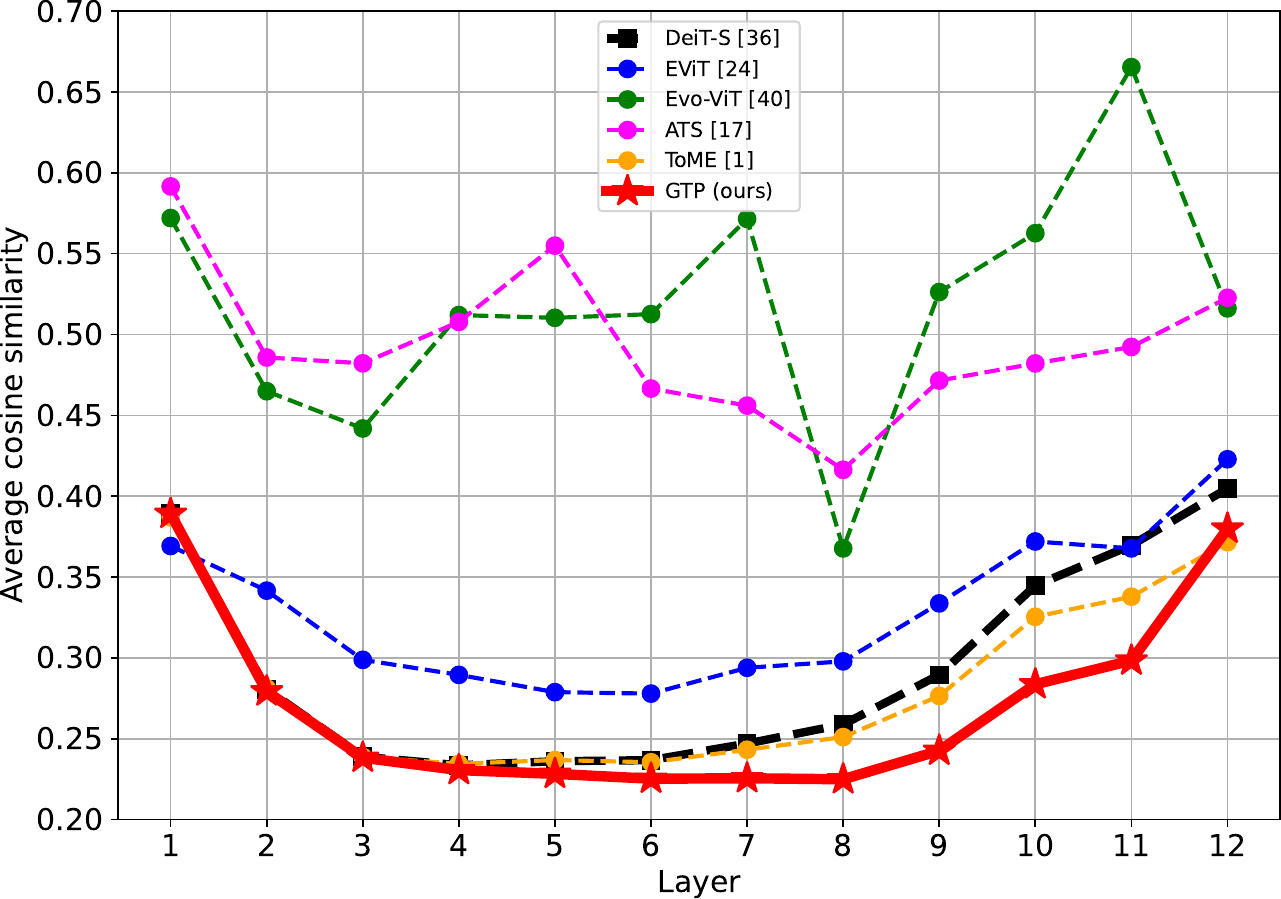} \vspace{-0.3em} 
    }
    \hspace{0.02\columnwidth}
    \subcaptionbox{Approx. 3.0 GMACs}{
        \includegraphics[width=.46\columnwidth]{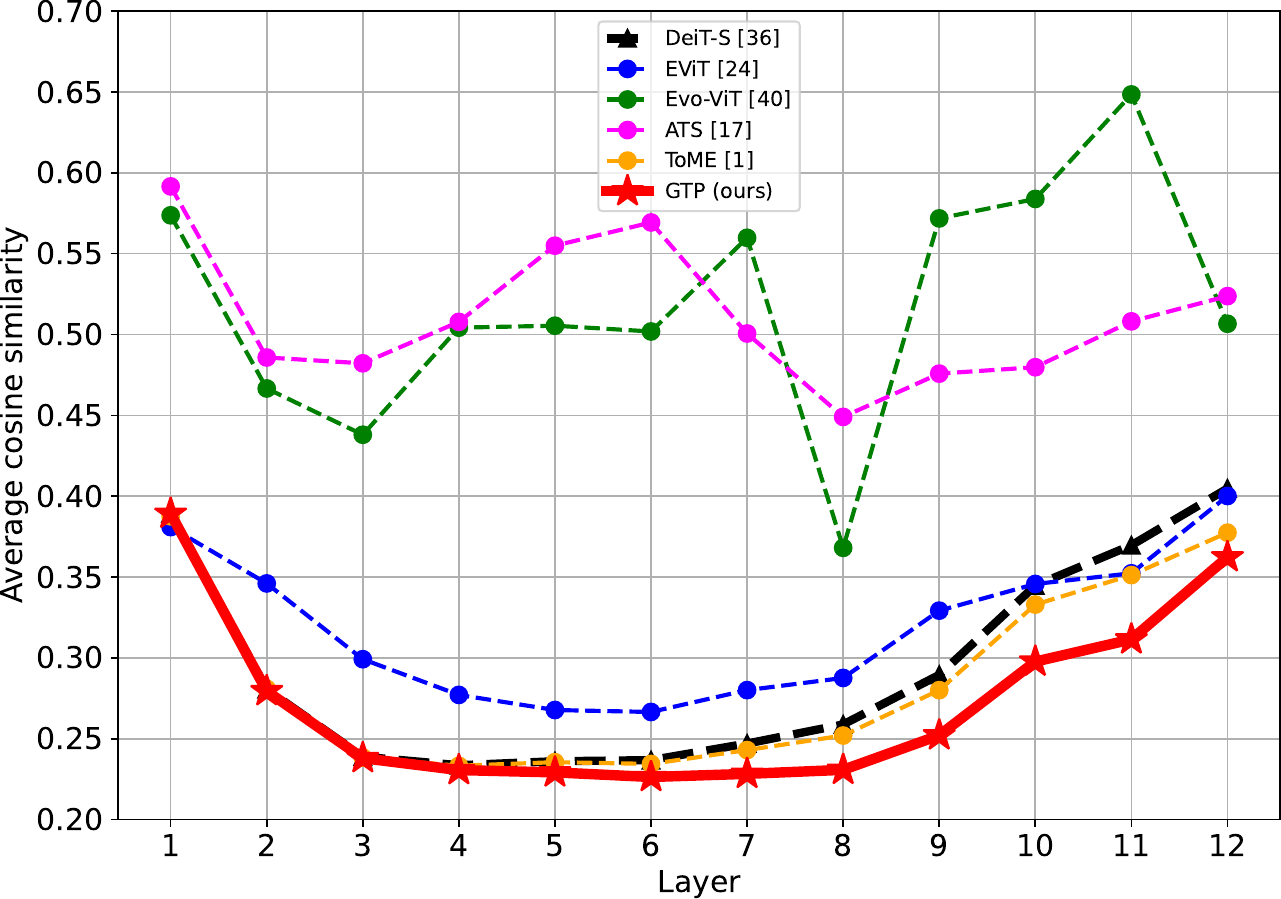} \vspace{-0.3em} 
    }
   \vspace{-0.5em}
   \caption{\textbf{Average cosine similarity.} We calculate the average cosine similarity among image tokens in each layer for various token reduction methods finetuned on DeiT-S.}
   \label{fig:antioversmoothing}
   \vspace{-0.8em}
\end{figure}

\vspace{-0.5em}
Figure \ref{fig:antioversmoothing} illustrates the trend of average cosine similarity between image tokens in each layer for various token reduction models. A higher average cosine similarity indicates a more severe oversmoothing problem where all the remaining tokens tend to be similar. Oversmoothing would lead to performance degradation both in GCN and ViT. Our GTP can mitigate the oversmoothing problem and yield lower similarities between image tokens, which is one of the key factors to our outstanding performance.

\vspace{-0.3em}
\subsection{Performance on ViT with more tokens}
\label{subsec:more tokens}
\vspace{-0.5em}
The token selection and fusion strategies of our GTP are more efficient than those of ToMe \cite{bolya2022token}. We provide a theoretical comparison of computational complexities in the supplementary material. In this section, we present empirical comparisons between GTP and ToMe, taking ViT-B-Patch8 \cite{dosovitskiy2020image} as the backbone. ViT-B-Patch8 contains 765 tokens, significantly more than ViT-B or DeiT-B which holds only 197 tokens. Experimental results in Table \ref{tab:more token results} demonstrate that with more tokens in the backbone ViT, our GTP achieves better performance and around 10\% faster inference speeds.

\vspace{-0.3em}
\subsection{Performance on ViTs without the [CLS] token}
\label{subsec:no [CLS] tokens}
\vspace{-0.5em}
As introduced in section \ref{subsubsec:token selection}, GTP selects tokens without the need for [CLS] token. In Table \ref{tab:noclstoken}, we present GTP's results on ViT-Medium-GAP \cite{dosovitskiy2020image} where global average pooling is used instead of the [CLS] token. It is worth noting that \cite{liang2021evit,xu2022evo,fayyaz2022adaptive,kong2022peeling} cannot work with this backbone.
\noindent

\vspace{-0.3em}
\subsection{Additional experiments}
\vspace{-0.5em}
In the appendix, we provide additional ablation studies on the hyperparameter $\alpha$ in token propagation and the number of semantic neighbours $M$. We also provide GTP's performance on large ViT models and a theoretical analysis of the computational complexity.

\vspace{-0.5em}
\section{Conclusion}
\vspace{-0.5em}
In this work, we treat the challenge of expediting ViTs for computing resource-constrained environments by reducing tokens as a token summarization task. We present a graph-based token propagation (GTP) method to address this issue without the need for finetuning. GTP constructs a sparse graph representation for image tokens and strategically selects less important tokens for propagation based on their regeneration difficulty and broadcasting ability. Then, GTP propagates the information of insignificant tokens into the other remaining tokens, which constitutes a condensed token representation. Extensive experiments have demonstrated the efficacy and efficiency of GTP. We hope our work can inspire future research in token reduction for ViTs.


{\small
\bibliographystyle{ieee_fullname}
\bibliography{egbib}

\begin{thebibliography}{10}\itemsep=-1pt

\bibitem{bolya2022token}
Daniel Bolya, Cheng-Yang Fu, Xiaoliang Dai, Peizhao Zhang, Christoph Feichtenhofer, and Judy Hoffman.
\newblock Token merging: Your vit but faster.
\newblock In {\em ICLR}, 2023.

\bibitem{bronstein2021geometric}
Michael~M Bronstein, Joan Bruna, Taco Cohen, and Petar Veli{\v{c}}kovi{\'c}.
\newblock Geometric deep learning: Grids, groups, graphs, geodesics, and gauges.
\newblock {\em arXiv preprint arXiv:2104.13478}, 2021.

\bibitem{chen2021crossvit}
Chun-Fu Chen, Quanfu Fan, and Rameswar Panda.
\newblock Crossvit: Cross-attention multi-scale vision transformer for image classification.
\newblock In {\em ICCV}, 2021.

\bibitem{chen2021regionvit}
Chun-Fu Chen, Rameswar Panda, and Quanfu Fan.
\newblock Regionvit: Regional-to-local attention for vision transformers.
\newblock In {\em ICLR}, 2022.

\bibitem{chen2021mobile}
Yinpeng Chen, Xiyang Dai, Dongdong Chen, Mengchen Liu, Xiaoyi Dong, Lu Yuan, and Zicheng Liu.
\newblock Mobile-former: Bridging mobilenet and transformer.
\newblock In {\em CVPR}, 2022.

\bibitem{chen2022vision}
Zhe Chen, Yuchen Duan, Wenhai Wang, Junjun He, Tong Lu, Jifeng Dai, and Yu Qiao.
\newblock Vision transformer adapter for dense predictions.
\newblock In {\em ICLR}, 2023.

\bibitem{chen2021visformer}
Zhengsu Chen, Lingxi Xie, Jianwei Niu, Xuefeng Liu, Longhui Wei, and Qi Tian.
\newblock Visformer: The vision-friendly transformer.
\newblock In {\em ICCV}, 2021.

\bibitem{chen2021dpt}
Zhiyang Chen, Yousong Zhu, Chaoyang Zhao, Guosheng Hu, Wei Zeng, Jinqiao Wang, and Ming Tang.
\newblock Dpt: Deformable patch-based transformer for visual recognition.
\newblock In {\em ACMMM}, 2021.

\bibitem{chu2021twins}
Xiangxiang Chu, Zhi Tian, Yuqing Wang, Bo Zhang, Haibing Ren, Xiaolin Wei, Huaxia Xia, and Chunhua Shen.
\newblock Twins: Revisiting the design of spatial attention in vision transformers.
\newblock In {\em NeurIPS}, 2021.

\bibitem{chuanyang2022savit}
Zheng Chuanyang, Zheyang Li, Kai Zhang, Zhi Yang, Wenming Tan, Jun Xiao, Ye Ren, and Shiliang Pu.
\newblock Savit: Structure-aware vision transformer pruning via collaborative optimization.
\newblock In {\em NeurIPS}, 2022.

\bibitem{dai2021coatnet}
Zihang Dai, Hanxiao Liu, Quoc~V Le, and Mingxing Tan.
\newblock Coatnet: Marrying convolution and attention for all data sizes.
\newblock In {\em NeurIPS}, 2021.

\bibitem{deng2009imagenet}
Jia Deng, Wei Dong, Richard Socher, Li-Jia Li, Kai Li, and Li Fei-Fei.
\newblock Imagenet: A large-scale hierarchical image database.
\newblock In {\em CVPR}, 2009.

\bibitem{ding2022davit}
Mingyu Ding, Bin Xiao, Noel Codella, Ping Luo, Jingdong Wang, and Lu Yuan.
\newblock Davit: Dual attention vision transformers.
\newblock In {\em ECCV}, 2022.

\bibitem{dong2021cswin}
Xiaoyi Dong, Jianmin Bao, Dongdong Chen, Weiming Zhang, Nenghai Yu, Lu Yuan, Dong Chen, and Baining Guo.
\newblock Cswin transformer: A general vision transformer backbone with cross-shaped windows.
\newblock In {\em CVPR}, 2022.

\bibitem{dosovitskiy2020image}
Alexey Dosovitskiy, Lucas Beyer, Alexander Kolesnikov, Dirk Weissenborn, Xiaohua Zhai, Thomas Unterthiner, Mostafa Dehghani, Matthias Minderer, Georg Heigold, Sylvain Gelly, Jakob Uszkoreit, and Neil Houlsby.
\newblock An image is worth 16x16 words: Transformers for image recognition at scale.
\newblock In {\em ICLR}, 2021.

\bibitem{d2021convit}
St{\'e}phane d’Ascoli, Hugo Touvron, Matthew~L Leavitt, Ari~S Morcos, Giulio Biroli, and Levent Sagun.
\newblock Convit: Improving vision transformers with soft convolutional inductive biases.
\newblock In {\em ICML}, 2021.

\bibitem{fang2023eva}
Yuxin Fang, Wen Wang, Binhui Xie, Quan Sun, Ledell Wu, Xinggang Wang, Tiejun Huang, Xinlong Wang, and Yue Cao.
\newblock Eva: Exploring the limits of masked visual representation learning at scale.
\newblock In {\em CVPR}, 2023.

\bibitem{fayyaz2022adaptive}
Mohsen Fayyaz, Soroush~Abbasi Koohpayegani, Farnoush~Rezaei Jafari, Sunando Sengupta, Hamid Reza~Vaezi Joze, Eric Sommerlade, Hamed Pirsiavash, and J{\"u}rgen Gall.
\newblock Adaptive token sampling for efficient vision transformers.
\newblock In {\em ECCV}, 2022.

\bibitem{jiang2021all}
Zi-Hang Jiang, Qibin Hou, Li Yuan, Daquan Zhou, Yujun Shi, Xiaojie Jin, Anran Wang, and Jiashi Feng.
\newblock All tokens matter: Token labeling for training better vision transformers.
\newblock In {\em NeurIPS}, 2021.

\bibitem{kipf2016semi}
Thomas~N Kipf and Max Welling.
\newblock Semi-supervised classification with graph convolutional networks.
\newblock In {\em ICLR}, 2017.

\bibitem{kong2022spvit}
Zhenglun Kong, Peiyan Dong, Xiaolong Ma, Xin Meng, Wei Niu, Mengshu Sun, Xuan Shen, Geng Yuan, Bin Ren, Hao Tang, et~al.
\newblock Spvit: Enabling faster vision transformers via latency-aware soft token pruning.
\newblock In {\em ECCV}, 2022.

\bibitem{kong2022peeling}
Zhenglun Kong, Haoyu Ma, Geng Yuan, Mengshu Sun, Yanyue Xie, Peiyan Dong, Xin Meng, Xuan Shen, Hao Tang, Minghai Qin, et~al.
\newblock Peeling the onion: Hierarchical reduction of data redundancy for efficient vision transformer training.
\newblock In {\em AAAI}, 2023.

\bibitem{lan2023couplformer}
Hai Lan, Xihao Wang, Hao Shen, Peidong Liang, and Xian Wei.
\newblock Couplformer: Rethinking vision transformer with coupling attention.
\newblock In {\em WACV}, 2023.

\bibitem{lefevre2010grass}
Kristen LeFevre and Evimaria Terzi.
\newblock Grass: Graph structure summarization.
\newblock In {\em ICDM}, 2010.

\bibitem{liang2021evit}
Youwei Liang, GE Chongjian, Zhan Tong, Yibing Song, Jue Wang, and Pengtao Xie.
\newblock Evit: Expediting vision transformers via token reorganizations.
\newblock In {\em ICLR}, 2021.

\bibitem{liu2023patchdropout}
Yue Liu, Christos Matsoukas, Fredrik Strand, Hossein Azizpour, and Kevin Smith.
\newblock Patchdropout: Economizing vision transformers using patch dropout.
\newblock In {\em WACV}, 2023.

\bibitem{liu2018graph}
Yike Liu, Tara Safavi, Abhilash Dighe, and Danai Koutra.
\newblock Graph summarization methods and applications: A survey.
\newblock {\em CSUR}, 2018.

\bibitem{liu2022swin}
Ze Liu, Han Hu, Yutong Lin, Zhuliang Yao, Zhenda Xie, Yixuan Wei, Jia Ning, Yue Cao, Zheng Zhang, Li Dong, Furu Wei, and Baining Guo.
\newblock Swin transformer v2: Scaling up capacity and resolution.
\newblock In {\em CVPR}, 2022.

\bibitem{liu2021swin}
Ze Liu, Yutong Lin, Yue Cao, Han Hu, Yixuan Wei, Zheng Zhang, Stephen Lin, and Baining Guo.
\newblock Swin transformer: Hierarchical vision transformer using shifted windows.
\newblock In {\em ICCV}, 2021.

\bibitem{lu2021soft}
Jiachen Lu, Jinghan Yao, Junge Zhang, Xiatian Zhu, Hang Xu, Weiguo Gao, Chunjing Xu, Tao Xiang, and Li Zhang.
\newblock Soft: Softmax-free transformer with linear complexity.
\newblock In {\em NeurIPS}, 2021.

\bibitem{marin2021token}
Dmitrii Marin, Jen-Hao~Rick Chang, Anurag Ranjan, Anish Prabhu, Mohammad Rastegari, and Oncel Tuzel.
\newblock Token pooling in vision transformers.
\newblock In {\em WACV}, 2021.

\bibitem{meng2022adavit}
Lingchen Meng, Hengduo Li, Bor-Chun Chen, Shiyi Lan, Zuxuan Wu, Yu-Gang Jiang, and Ser-Nam Lim.
\newblock Adavit: Adaptive vision transformers for efficient image recognition.
\newblock In {\em CVPR}, 2022.

\bibitem{torch_sparse}
PyTorch.
\newblock Torch.sparse, pytorch 2.0 documentation.
\newblock \url{https://pytorch.org/docs/stable/sparse.html}, 2023.

\bibitem{rao2021dynamicvit}
Yongming Rao, Wenliang Zhao, Benlin Liu, Jiwen Lu, Jie Zhou, and Cho-Jui Hsieh.
\newblock Dynamicvit: Efficient vision transformers with dynamic token sparsification.
\newblock In {\em NeurIPS}, 2021.

\bibitem{riondato2017graph}
Matteo Riondato, David Garc{\'\i}a-Soriano, and Francesco Bonchi.
\newblock Graph summarization with quality guarantees.
\newblock {\em DMKD}, 2017.

\bibitem{ryoo2021tokenlearner}
Michael Ryoo, AJ Piergiovanni, Anurag Arnab, Mostafa Dehghani, and Anelia Angelova.
\newblock Tokenlearner: Adaptive space-time tokenization for videos.
\newblock In {\em NeurIPS}, 2021.

\bibitem{srinivas2021bottleneck}
Aravind Srinivas, Tsung-Yi Lin, Niki Parmar, Jonathon Shlens, Pieter Abbeel, and Ashish Vaswani.
\newblock Bottleneck transformers for visual recognition.
\newblock In {\em CVPR}, 2021.

\bibitem{touvron2021training}
Hugo Touvron, Matthieu Cord, Matthijs Douze, Francisco Massa, Alexandre Sablayrolles, and Herv{\'e} J{\'e}gou.
\newblock Training data-efficient image transformers \& distillation through attention.
\newblock In {\em ICLR}, 2021.

\bibitem{velivckovic2017graph}
Petar Veli{\v{c}}kovi{\'c}, Guillem Cucurull, Arantxa Casanova, Adriana Romero, Pietro Lio, and Yoshua Bengio.
\newblock Graph attention networks.
\newblock In {\em ICLR}, 2018.

\bibitem{wang2021pyramid}
Wenhai Wang, Enze Xie, Xiang Li, Deng-Ping Fan, Kaitao Song, Ding Liang, Tong Lu, Ping Luo, and Ling Shao.
\newblock Pyramid vision transformer: A versatile backbone for dense prediction without convolutions.
\newblock In {\em ICCV}, 2021.

\bibitem{wu2021cvt}
Haiping Wu, Bin Xiao, Noel Codella, Mengchen Liu, Xiyang Dai, Lu Yuan, and Lei Zhang.
\newblock Cvt: Introducing convolutions to vision transformers.
\newblock In {\em ICCV}, 2021.

\bibitem{xu2022evo}
Yifan Xu, Zhijie Zhang, Mengdan Zhang, Kekai Sheng, Ke Li, Weiming Dong, Liqing Zhang, Changsheng Xu, and Xing Sun.
\newblock Evo-vit: Slow-fast token evolution for dynamic vision transformer.
\newblock In {\em AAAI}, 2022.

\bibitem{yang2021focal}
Jianwei Yang, Chunyuan Li, Pengchuan Zhang, Xiyang Dai, Bin Xiao, Lu Yuan, and Jianfeng Gao.
\newblock Focal attention for long-range interactions in vision transformers.
\newblock In {\em NeurIPS}, 2021.

\bibitem{yoo2023enriched}
Jinsu Yoo, Taehoon Kim, Sihaeng Lee, Seung~Hwan Kim, Honglak Lee, and Tae~Hyun Kim.
\newblock Enriched cnn-transformer feature aggregation networks for super-resolution.
\newblock In {\em WACV}, 2023.

\bibitem{zhai2021scaling}
Xiaohua Zhai, Alexander Kolesnikov, Neil Houlsby, and Lucas Beyer.
\newblock Scaling vision transformers.
\newblock In {\em CVPR}, 2021.

\bibitem{zhang2021multi}
Pengchuan Zhang, Xiyang Dai, Jianwei Yang, Bin Xiao, Lu Yuan, Lei Zhang, and Jianfeng Gao.
\newblock Multi-scale vision longformer: A new vision transformer for high-resolution image encoding.
\newblock In {\em ICCV}, 2021.

\bibitem{zhou2021deepvit}
Daquan Zhou, Bingyi Kang, Xiaojie Jin, Linjie Yang, Xiaochen Lian, Zihang Jiang, Qibin Hou, and Jiashi Feng.
\newblock Deepvit: Towards deeper vision transformer.
\newblock {\em arXiv preprint arXiv:2103.11886}, 2021.

\bibitem{zong2022self}
Zhuofan Zong, Kunchang Li, Guanglu Song, Yali Wang, Yu Qiao, Biao Leng, and Yu Liu.
\newblock Self-slimmed vision transformer.
\newblock In {\em ECCV}, 2022.

\end{thebibliography}
}

\appendix

\section{Implementation optimizations} 
First, we note that both the spatial graph and semantic graph are sparse graphs, with graph sparsity $\frac{8}{N}$ and $\frac{M}{N}$, respectively. For ViT \cite{dosovitskiy2020image} and DeiT \cite{touvron2021training} models, the total number of image tokens $N$ is usually 196, which indicates less than 5\% non-trivial values in the two adjacency matrices. As a result, the mixed graph is also a sparse graph whose sparsity is no more than $\frac{8+M}{N}$ (less than 7\% on average). Therefore, we can store the graph in sparse tensors \cite{torch_sparse} and perform sparse matrix multiplication to accelerate the graph propagation in Sections 3.2.2 and 3.2.3. Besides, for batched inputs that the sparse matrix multiplication does not support, we can use the scatter reduction operation to avoid the dense matrix multiplication. Second, the threshold $T_i$ for sparsifying the semantic graph can be determined by finding the $M^{\text{th}}$ largest value in the unsorted array $\mathcal{A}^{\text{semantic}}_{i}$ with a complexity $O(N)$ rather than sorting the whole array with a complexity $O(N\log N)$.

\section{Experiment settings} 
We provide the hyperparameter settings for the compared methods in Tables \ref{tab:hyperparameter for deit-s} and \ref{tab:hyperparameter for deit-b}. These hyperparameters are used to control the reduced computational complexity for the backbone ViT, ensuring fair comparisons.
\begin{table*}
  \centering
  \footnotesize
  \setlength{\tabcolsep}{4pt}
  \begin{tabular}{|l|l|l|l|l|l|l|}
    \hline
    \multirow{2}{*}{Backbone} & \multirow{2}{*}{Method} & \multirow{2}{*}{Hyperparameter} & \multicolumn{4}{c|}{Approximate computational complexity}\\
    \cline{4-7}
    & & & 3.5 GMACs & 3.0 GMACs & 2.6 GMACs & 2.3 GMACs \\
    \hline
    \multirow{7}{*}{DeiT-S \cite{touvron2021training}} & DynamicViT \cite{rao2021dynamicvit} & base keep ratio $\rho$ & $\rho=0.8$ & $\rho=0.7$ & $\rho=0.6$ & $\rho=0.5$ \\
    & EViT \cite{liang2021evit} & token keeping rate $k$ & $k=0.8$ & $k=0.7$ & $k=0.6$ & $k=0.5$ \\
    & Evo-ViT \cite{xu2022evo} & selection ratio $p$ & $p=0.7$ & $p=0.5$ & $p=0.4$ & $p=0.3$  \\
    & Tri-Level \cite{kong2022peeling} & token keep ratio $R_T$ & $R_T=0.8$ & $R_T=0.7$ & $R_T=0.6$ & $R_T=0.5$\\
    & ToMe \cite{bolya2022token} & token reduce $r$ per layer & $r=8$ & $r=11$ & $r=14$ & $r=16$\\
    & ATS \cite{fayyaz2022adaptive} & ATS block layers & layer 7 to 11 & layer 6 to 11 & layer 5 to 11 & layer 3 to 11 \\
    & GTP (ours) & propagated tokens $P$ per layer & $P=8$ & $P=11$ & $P=14$ & $P=16$ \\
    \hline
  \end{tabular}
  \caption{\small\textbf{Hyperparameter settings for the baseline methods, taking DeiT-S as the backbone.}}
  \label{tab:hyperparameter for deit-s}
\end{table*}
\begin{table*}
  \centering
  \footnotesize
  \setlength{\tabcolsep}{4pt}
  \begin{tabular}{|l|l|l|l|l|l|l|l|}
    \hline
    \multirow{2}{*}{Backbone} & \multirow{2}{*}{Method} & \multirow{2}{*}{Hyperparameter} & \multicolumn{5}{c|}{Approximate computational complexity}\\
    \cline{4-8}
    & & & 15.3 GMACs & 13.1 GMACs & 11.6 GMACs & 9.8 GMACs & 8.8 GMACs\\
    \hline
    \multirow{7}{*}{DeiT-B \cite{touvron2021training}} & DynamicViT \cite{rao2021dynamicvit} & base keep ratio $\rho$ & $\rho=0.9$ & $\rho=0.8$ & $\rho=0.7$ & $\rho=0.6$ & $\rho=0.5$ \\
    & EViT \cite{liang2021evit} & token keeping rate $k$ & $k=0.9$ & $k=0.8$ & $k=0.7$ & $k=0.6$ & $k=0.5$ \\
    & Evo-ViT \cite{xu2022evo} & selection ratio $p$ & $p=0.8$ & $p=0.7$ & $p=0.5$ & $p=0.4$ & $p=0.3$  \\
    & Tri-Level \cite{kong2022peeling} & token keep ratio $R_T$ & $R_T=0.9$ & $R_T=0.8$ & $R_T=0.7$ & $R_T=0.6$ & $R_T=0.5$\\
    & ToMe \cite{bolya2022token} & token reduce $r$ per layer & $r=4$ & $r=8$ & $r=11$ & $r=14$ & $r=16$\\
    & ATS \cite{fayyaz2022adaptive} & ATS block layers & layer 9 to 11 & layer 8 to 11 & layer 6 to 11 & layer 3 to 11 & layer 1 to 11 \\
    & GTP (ours) & propagated tokens $P$ per layer & $P=4$ & $P=8$ & $P=11$ & $P=14$ & $P=16$ \\
    \hline
  \end{tabular}
  \caption{\small\textbf{Hyperparameter settings for the baseline methods, taking DeiT-B as the backbone.}}
  \label{tab:hyperparameter for deit-b}
\end{table*}

\section{Additional experiments}
\subsection{Larger ViT backbones}
In the main submission, we have validated GTP's effectiveness on small to medium-sized ViT backbones. Moreover, we employ GTP on two large-size ViT backbones: ViT-L \cite{dosovitskiy2020image} and EVA-L \cite{fang2023eva}, which represent ViT backbones with and without the [CLS] token, respectively. Since ViT-L and EVA-L both have 24 layers, the maximum number of propagated tokens $P$ per layer is limited to 8. We also reproduce the state-of-the-art ToMe \cite{bolya2022token} on these models and present the comparisons in Table \ref{tab:larger model results}. It is worth noting that ToMe consumes considerable GPU memory for computing the cosine similarity in each layer. We omit the evaluation on larger models (-H, -G) due to hardware constraints.

From Table \ref{tab:larger model results}, we point out that our GTP outperforms ToMe in both model performance and efficiency. Such performance difference is even more significant on ViT backbones without the [CLS] token. For instance, GTP achieves 85.4\% top-1 accuracy at 212.0 image/second when taking EVA-L as the backbone, \textbf{surpassing ToMe's 80.1\% top-1 accuracy by a significant 5.3\%} at a similar inference speed.
\begin{table*}
  \footnotesize
  \setlength{\tabcolsep}{2pt}
  \centering
  \begin{tabular}{|l|l|cc|cc|cc|cc|cc|cc|cc|cc|cc|}
    \hline
    \multirow{3}{*}{Backbone} & \multirow{3}{*}{Method} & \multicolumn{2}{c|}{\tiny $P=0$} & \multicolumn{2}{c|}{\tiny $P=1$} & \multicolumn{2}{c|}{\tiny $P=2$} & \multicolumn{2}{c|}{\tiny $P=3$} & \multicolumn{2}{c|}{\tiny $P=4$} & \multicolumn{2}{c|}{\tiny $P=5$} & \multicolumn{2}{c|}{\tiny $P=6$} & \multicolumn{2}{c|}{\tiny $P=7$} & \multicolumn{2}{c|}{\tiny $P=8$}\\
    \cline{3-20}
    & & \makecell{Acc.\\(\%)} & \makecell{Speed\\(img/s)} & \makecell{Acc.\\(\%)} & \makecell{Speed\\(img/s)} & \makecell{Acc.\\(\%)} & \makecell{Speed\\(img/s)} & \makecell{Acc.\\(\%)} & \makecell{Speed\\(img/s)} & \makecell{Acc.\\(\%)} & \makecell{Speed\\(img/s)} & \makecell{Acc.\\(\%)} & \makecell{Speed\\(img/s)} & \makecell{Acc.\\(\%)} & \makecell{Speed\\(img/s)} & \makecell{Acc.\\(\%)} & \makecell{Speed\\(img/s)} & \makecell{Acc.\\(\%)} & \makecell{Speed\\(img/s)}\\
    \hline
    \multirow{2}{*}{ViT-L \cite{touvron2021training}} & ToMe \cite{bolya2022token} & \multirow{2}{*}{85.8} & \multirow{2}{*}{123.4} & \textbf{85.8} & 122.0 & 85.7 & 132.1 & 85.5 & 142.2 & 85.3 & 153.8 & 85.0 & 169.0 & 84.7 & 184.8 & 84.2 & 204.0 & \textbf{83.7} & 228.5\\
    & GTP (ours) & & & \textbf{85.8} & \textbf{125.4} & \textbf{85.8} & \textbf{134.4} & \textbf{85.8} & \textbf{144.9} & \textbf{85.5} & \textbf{157.4} & \textbf{85.3} & \textbf{172.0} & \textbf{85.0} & \textbf{188.3} & \textbf{84.3} & \textbf{208.8} & \textbf{83.7} & \textbf{234.0} \\
    \hline
    \multirow{2}{*}{EVA-L \cite{fang2023eva}} & ToMe \cite{bolya2022token} & \multirow{2}{*}{87.9} & \multirow{2}{*}{123.1} & 87.7 & 123.6 & 87.6 & 132.3 & 87.3 & 142.5 & 87.0 & 155.2 & 86.5 & 169.6 & 85.6 & 185.9 & 84.0 & 207.0 & 80.1 & 230.9\\
    & GTP (ours) & & & \textbf{87.9} & \textbf{125.6} & \textbf{87.8} & \textbf{134.6} & \textbf{87.8} & \textbf{145.3} & \textbf{87.7} & \textbf{158.0} & \textbf{87.5} & \textbf{172.5} & \textbf{87.2} & \textbf{188.9} & \textbf{86.5} & \textbf{212.0} & \textbf{85.4} & \textbf{234.9} \\
    \hline
  \end{tabular}
  \caption{\small\textbf{Performance on larger ViT models.} We validate GTP's performance on two large-size ViT models, ViT-L \cite{dosovitskiy2020image} and EVA-L \cite{fang2023eva}, where ViT-L employs the [CLS] token while EVA-L does not have the [CLS] token. Since both models have 24 layers, we can eliminate at most 8 tokens per layer. Bond font means better. GTP constantly outperform ToMe on large ViT models with and without the [CLS] token.}
  \label{tab:larger model results}
\end{table*}

\subsection{Graph propagation hyperparameter $\alpha$}
\begin{figure}
    \subcaptionbox{DeiT-S}{
        \includegraphics[width=.46\columnwidth, trim=0.7cm 0 0 0]{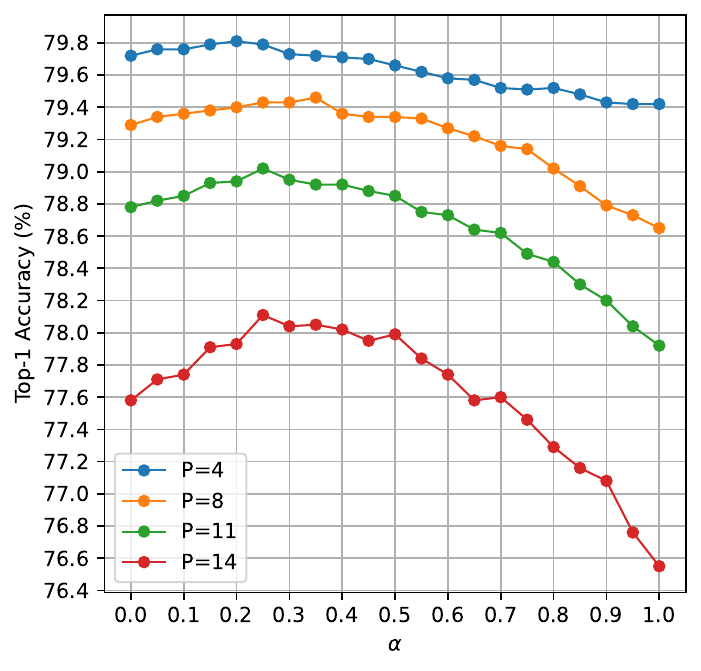}
    }
    \hspace{0.01\columnwidth}
    \subcaptionbox{DeiT-B}{
        \includegraphics[width=.46\columnwidth, trim=0.7cm 0 0 0]{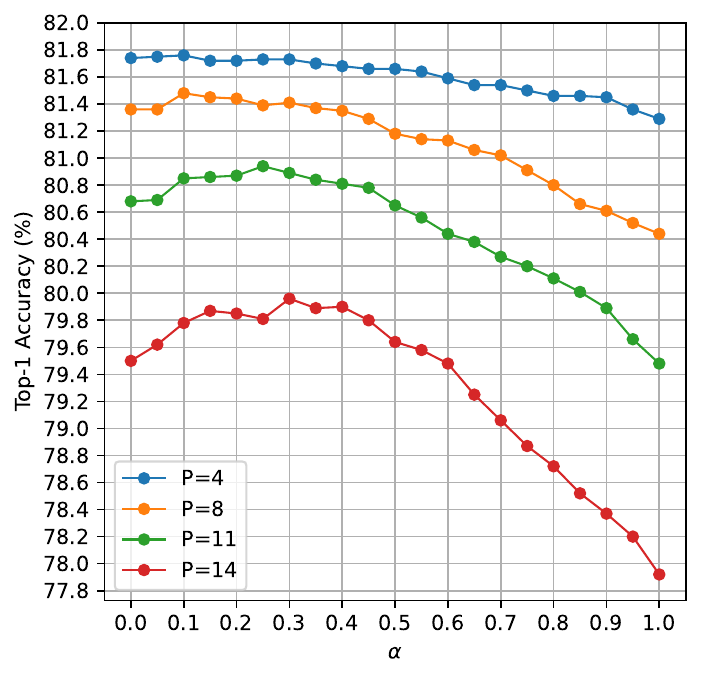}
    }
    \caption{\textbf{Top-1 accuracy of GTP on ImageNet-1K \cite{deng2009imagenet} for various $\alpha$s.} We evaluate the performance of GTP on both DeiT-S and DeiT-B \cite{touvron2021training} \textit{without finetuning} w.r.t. different $\alpha$. For fair comparisons, we employ the same graph type for propagation and set the attention sparsity static at 0.6 for DeiT-S and 0.5 for DeiT-B, respectively. The findings in this experiment are consistent across different settings.}
    \label{fig:alpha study}
\end{figure}
In the graph summarization process $\mathbi{X}^s= \mathbi{X}^k + \alpha\hat{\mathcal{A}}^{p}\mathbi{X}^p$, we use $\alpha$ to control the magnitude of information broadcast from propagated tokens $\mathbi{X}^p$ to kept tokens $\mathbi{X}^k$. In this section, we investigate the performance of GTP with respect to different $\alpha$ and visualize the results in Figure \ref{fig:alpha study}. In general, as $\alpha$ increases within the range of $[0, 1]$, the corresponding accuracy first rises and then declines, reaching its peak between 0.2\textasciitilde0.4 for DeiT-S and 0.1\textasciitilde0.3 for DeiT-B. We explain this phenomenon from two aspects. First, when $\alpha$ is approaching 0, the propagated information becomes trivial, leading to a situation where the propagated tokens' information is not preserved. When $\alpha=0$, this process merely prunes tokens. Secondly, as $\alpha$ increases, the propagated information gradually dominates the original information of the remaining tokens, which results in an over-smoothing problem and subsequently hinders performance.

\subsection{The number of graph neighbours}
\begin{figure*}
    \subcaptionbox{\scriptsize GTP on DeiT-S with semantic graph}{
        \includegraphics[width=.47\columnwidth, trim=0.7cm 0 0 0]{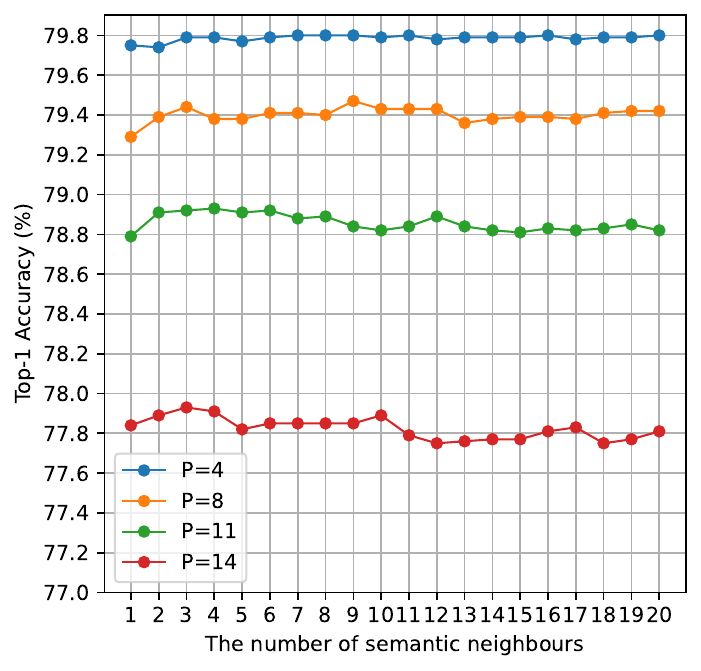}
    }
    \hspace{0.02\columnwidth}
    \subcaptionbox{\scriptsize GTP on DeiT-S with mixed graph}{
        \includegraphics[width=.47\columnwidth, trim=0.7cm 0 0 0]{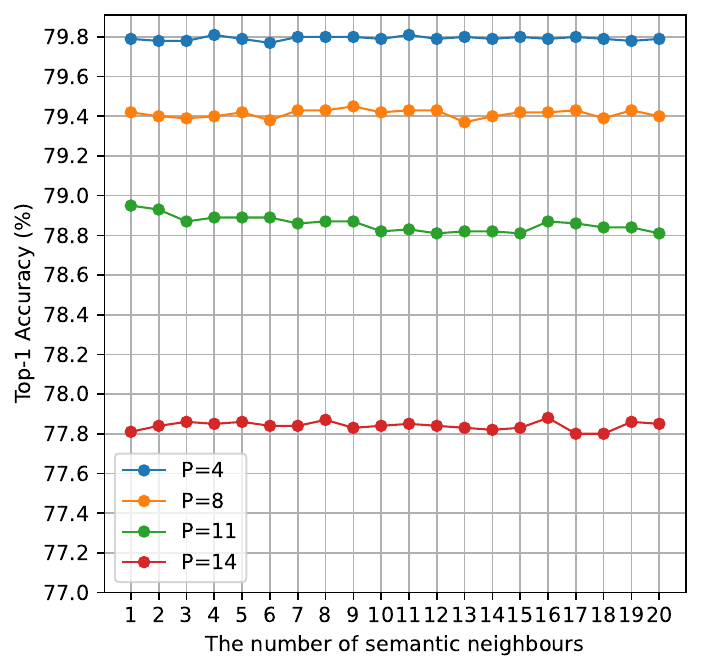}
    }
    \hspace{0.02\columnwidth}
    \subcaptionbox{\scriptsize GTP on DeiT-B with semantic graph}{
        \includegraphics[width=.47\columnwidth, trim=0.7cm 0 0 0]{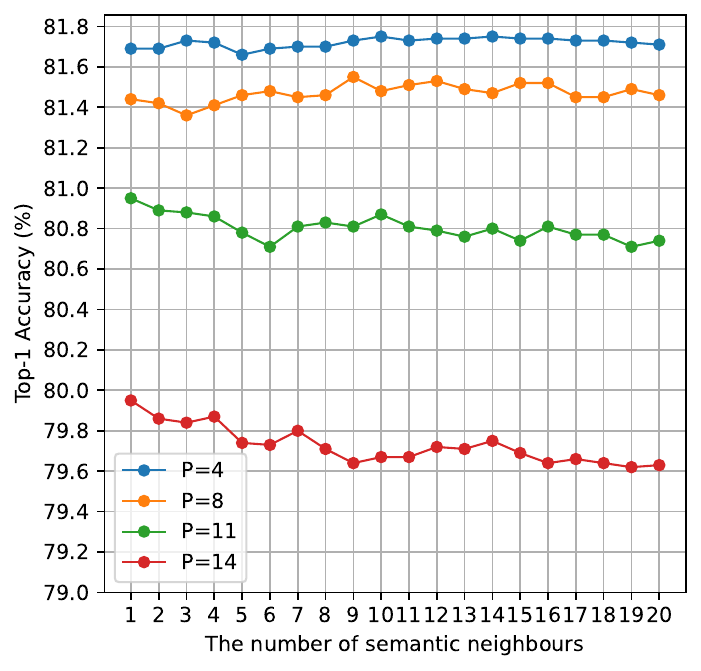}
     }
    \hspace{0.02\columnwidth}
    \subcaptionbox{\scriptsize GTP on DeiT-B with mixed graph}{
        \includegraphics[width=.47\columnwidth, trim=0.7cm 0 0 0]{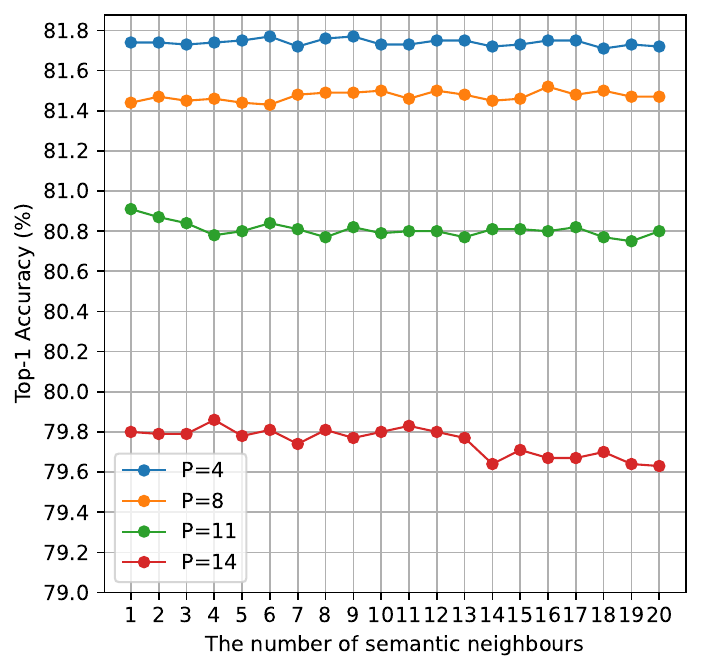}
    }
    \caption{\textbf{Top-1 accuracy of GTP with various numbers of semantic neighbours $M$.} We evaluate the performance of GTP on both DeiT-S and DeiT-B \cite{touvron2021training} \textit{without finetuning} w.r.t. different numbers of semantically connected neighbours.}
    \label{fig:number of semantic neighbours study}
\end{figure*}
We study the influence of the number of semantic neighbours $M$ on model performance and plot the accuracy in Figure \ref{fig:number of semantic neighbours study}. For fair comparisons, we apply GTP on DeiT-S and DeiT-B with static attention sparsity and alpha at 0.5 and 0.2, respectively. Figures \ref{fig:number of semantic neighbours study}(a) and \ref{fig:number of semantic neighbours study}(c) illustrate the results obtained by only employing the semantic graph for token propagation, with respect to different $M$. It can be observed that when the number of propagated tokens $P$ is small (e.g., $P=4$ or $P=8$), increasing the semantic neighbours would first slightly improves the accuracy and then converges. However, when $P$ becomes large (e.g., $P=14$), increasing the semantic neighbours may lead to a performance drop. This can be attributed to the aggregation of redundant information, where one kept token encapsulates an excessive number of propagated tokens that may not be semantically close to it. Figures \ref{fig:number of semantic neighbours study}(b) and \ref{fig:number of semantic neighbours study}(d) show that integrating the semantic graph with the spatial graph stabilizes the trend of accuracy, indicating the significance of the spatial relationship in token summarization.

\subsection{Computational complexity comparison}
\label{complexity}
As stated in the Introduction section, ToMe \cite{bolya2022token} encounters a computational bottleneck in the pair-wise token matching process, whose computational complexity is proportional to the feature dimensions and the square of the number of tokens. Compared with ToMe, GTP demonstrates faster inference speed and accomplishes better information preservation results. In this section, we provide an in-depth analysis of the enhancements in computational efficiency achieved by GTP.

As a plug-and-play component, GTP inserts the token summarization module between the MHSA layer and the FFN layer in each ViT block, which behaves analogously to ToMe. Therefore, when the number of eliminated tokens is the same, the computational complexity of the backbone model for GTP and ToMe should be the same. Consequently, we only consider the additional computational costs (e.g., token matching, token selection and token propagation) introduced by the two models in this analysis. We list the denotations before the theoretical analysis as follows:
\begin{equation}
    \begin{aligned}
        N :& \text{ The total number of tokens in the backbone network.} \\
        N_l :& \text{ The number of remaining tokens in layer $l$, where}\\ 
        & \,\, \text{$N_l=N-(l-1)M$}. \\
        M :& \text{ The number of eliminated tokens in each layer.} \\
        C :& \text{ The dimension of features.} \\
        L :& \text{ The total number of layers.} \\
        H :& \text{ The number of heads.}
    \end{aligned}
\end{equation}
For ToMe, the token matching processing first splits tokens into two sets and then calculates the cosine similarity between each pair of tokens from the two sets. The computational complexity for this process in layer $l$ is $\frac{1}{4}N_{l}^2C$. Besides, ToMe merges $M$ tokens in each layer, whose total computational complexity is $MC$ in each layer. As a result, the total additional computational complexity $G_{\text{ToMe}}$ introduced by ToMe is calculated as
\begin{equation}
    \begin{aligned}
        G_{\text{ToMe}} &= \sum_{l=1}^{L}{({\frac{1}{4}N_{l}^2C} + MC)} \\
        &= \frac{1}{4}C\sum_{l=1}^{L}{N_{l}^2} + LMC\\
        &=\frac{1}{4}C\sum_{l=1}^{L}{(N-(l-1)M)^2} + LMC\\
        &=\frac{1}{4}C\sum_{l=1}^{L}{(N^2-2(l-1)NM+(l-1)^2M^2)}\\
        &\quad + LMC\\
        &=\frac{1}{4}LN^2C+\frac{1}{4}(L-L^2)NMC\\
        &\quad + (\frac{1}{12}L^3+\frac{1}{12}L^2+\frac{1}{8}L)M^2C + LMC
    \end{aligned}
\end{equation}

We then calculate the computational complexity for GTP. GTP first constructs the semantic graph for an input image after the token embedding layer with a computational complexity $N^2C$. Next, it selects tokens with a computational complexity at most $HN_l$ in layer $l$. And finally, the tokens are propagated with computational complexity at most $(N_l-M)MC$ in layer $l$. Consequently, the total additional computational complexity $G_{\text{GTP}}$ of GTP is
\begin{equation}
    \begin{aligned}
        G_{\text{GTP}} &= N^2C + \sum_{l=1}^{L}{(HN_l+(N_l-M)MC)}\\
        &= N^2C + \sum_{l=1}^{L}(H(N-(l-1)M)\\
        &\quad +(N-(l-1)M-M)MC) \\
        &= N^2C + LHN + LMNC - \frac{1}{2}(L^2-L)HM \\
        &\quad - \frac{1}{2}(L+L^2)M^2C. \\
    \end{aligned}
\end{equation}

\begin{figure}
    \subcaptionbox{}{
        \includegraphics[width=.47\columnwidth, trim=0.25cm 0 0 0]{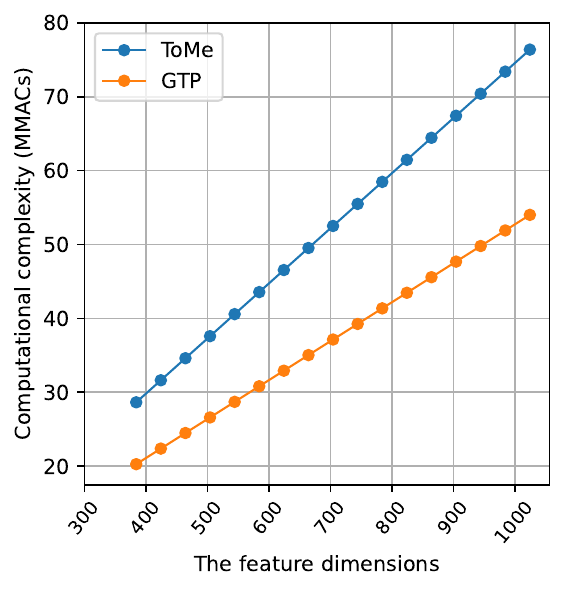}
    }
    \subcaptionbox{}{
        \includegraphics[width=.47\columnwidth, trim=0.7cm 0 0 0]{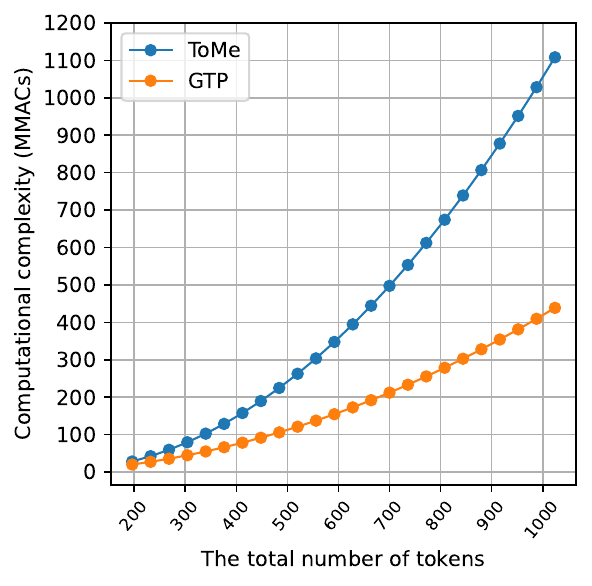}
    }
    \caption{\textbf{Comparisons on the additional computational complexities introduced by ToMe \cite{bolya2022token} and our GTP.} We plot the computational complexity (measured in MMACs) with respect to the dimension of token features in (a) and the total number of tokens in (b).}
    \label{fig:inference speed}
\end{figure}

Given $N=197, L=12, H=6, C=384$ and $M=8$ for DeiT-S, we can get $G_{\text{GTP}}\approx20.1$MMACs, which is smaller than $G_{\text{ToMe}}\approx28.3$MMACs. On DeiT-B where $N=197, L=12, M=8, H=12$ and $C=768$, we observe that $G_{\text{GTP}}\approx40.5$MMACs is much smaller than $G_{\text{ToMe}}\approx57.3$MMACs. Figure \ref{fig:inference speed} illustrates the additional computational complexity change with respect to the total number of tokens $N$ and the feature dimensions $C$. It is obvious that our GTP introduces far less additional computational complexity than ToMe, and the difference signifies when $N$ and $C$ increase. It indicates the efficiency of GTP on large-size ViTs whose feature dimensions may exceed 1024, as well as on ViTs for dense prediction tasks where the number of tokens may be more than 1024.

\end{document}